\documentclass{article}

\usepackage{microtype}
\usepackage{graphicx}
\usepackage{subfigure}
\usepackage{booktabs} 
\usepackage{enumitem}
\usepackage{bm}
\usepackage{enumitem}

\usepackage{hyperref}

\usepackage[accepted]{icml2025}

\usepackage{amsmath}
\usepackage{amssymb}
\usepackage{mathtools}
\usepackage{amsthm}

\usepackage[capitalize,noabbrev]{cleveref}

\theoremstyle{plain}
\newtheorem{theorem}{Theorem}[section]

\theoremstyle{definition}

\theoremstyle{remark}

\usepackage[textsize=tiny]{todonotes}

\newcommand{\nop}[1]{}

\icmltitlerunning{Beyond the Permutation Symmetry of Transformers: The Role of Rotation for Model Fusion}

\begin{document}

\twocolumn[
\icmltitle{Beyond the Permutation Symmetry of Transformers:\\The Role of Rotation for Model Fusion}



\icmlsetsymbol{equal}{*}

\begin{icmlauthorlist}
\icmlauthor{Binchi Zhang}{equal,uva}
\icmlauthor{Zaiyi Zheng}{equal,uva}
\icmlauthor{Zhengzhang Chen}{nec}
\icmlauthor{Jundong Li}{uva}
\end{icmlauthorlist}

\icmlaffiliation{uva}{University of Virginia}
\icmlaffiliation{nec}{NEC Laboratories America}

\icmlcorrespondingauthor{Jundong Li}{jundong@virginia.edu}

\icmlkeywords{Machine Learning, ICML}

\vskip 0.3in
]



\printAffiliationsAndNotice{\icmlEqualContribution} 

\begin{abstract}
Symmetry in the parameter space of deep neural networks (DNNs) has proven beneficial for various deep learning applications. 
A well-known example is the permutation symmetry in Multi-Layer Perceptrons (MLPs), where permuting the rows of weight matrices in one layer and applying the inverse permutation to adjacent layers yields a functionally equivalent model. 
While permutation symmetry fully characterizes the equivalence set for MLPs, its discrete nature limits its utility for transformers. 
In this paper, we introduce rotation symmetry, a novel form of parameter space symmetry for transformers that generalizes permutation symmetry by rotating parameter matrices in self-attention layers. 
Unlike permutation symmetry, rotation symmetry operates in a continuous domain, thereby significantly expanding the equivalence set for transformers. 
Based on this property, we propose a theoretically optimal parameter matching algorithm as a plug-and-play module to enhance model fusion. 
We evaluate our approach using pre-trained transformers across diverse natural language and vision tasks. 
Experimental results demonstrate that our rotation symmetry-based matching algorithm substantially improves model fusion, highlighting the potential of parameter space symmetry to facilitate model fusion. 
Our code is available on 
\url{https://github.com/zhengzaiyi/RotationSymmetry}.
\end{abstract}

\section{Introduction}
Parameter space symmetry is an intriguing property of neural networks that has garnered increasing attention in recent years~\citep{du2018algorithmic,armenta2021representation,kunin2021neural,simsek2021geometry,entezari2022role,grigsby2023hidden,lim2024empirical,ziyin2024parameter,ziyin2025remove}. 
One of the most studied forms of parameter space symmetry is permutation symmetry~\citep{ainsworth2023git,entezari2022role}. 
For instance, in a two-layer MLP, permuting the rows of the weight matrix in the first layer and applying the corresponding inverse permutation to the second layer results in a functionally equivalent model, \textit{i.e.}, the outputs of the original and permuted models remain identical for any given input~\citep{ainsworth2023git}.
All functionally equivalent models corresponding to weight permutations form an equivalence set, which provides theoretical insights into neural network optimization, such as the linear mode connectivity of loss landscapes~\citep{entezari2022role,zhou2023going,ferbach2024proving}. 
In addition, permutation symmetry has also proven helpful in advancing neural network applications, such as model fusion~\citep{singh2020model,ainsworth2023git} and optimization~\citep{zhao2024improving,zamir2025improving}.

Although parameter space symmetry has been extensively studied in classical neural network architectures, such as MLPs and CNNs, the understanding of its application in transformers~\citep{vaswani2017attention} remains limited. 
Transformers have seen rapid advancements in recent years, achieving remarkable success in a wide range of applications~\citep{yun2019graph,lewis2020bart,raffel2020exploring,clark2020electra,zhou2021informer,he2024explaining,zhu2024understanding,zheng2024kg,zhang2025resolving,wei2025large}.
The transformer architecture is built upon two primary submodules: feedforward networks and self-attention layers. 
The feedforward network, which is structurally similar to MLPs, naturally inherits the permutation symmetry that has been extensively studied in the existing literature. 
Self-attention layers, on the other hand, involve a unique attention mechanism powered by matrix products of queries, keys, and values, which introduce additional potentials for symmetry beyond permutations. 
Permutation symmetries limit the equivalence set of neural networks to \textit{discrete operations}, which aligns well with MLPs due to their element-wise activations (\textit{e.g.}, ReLU~\citep{glorot2011deep}). 
In contrast, \textit{the continuous nature} of the matrix operations in self-attention layers necessitates more flexible operations to fully characterize their equivalence set.

In this paper, we introduce \textbf{rotation symmetry}, a novel form of parameter space symmetry for self-attention layers in transformers. 
Specifically, 
we analyze the query and key matrices jointly and demonstrate that applying a rotation to the query matrix, followed by the corresponding inverse rotation to the key matrix, preserves the query-key product. 
Additionally, we find that the same rotation rule can be applied to the value and output matrices.
Our findings provide novel insights into the functional invariance of the attention mechanism and extend the permutation symmetry~\citep{singh2020model,wang2020federated,tatro2020optimizing,entezari2022role,ainsworth2023git,imfeld2024transformer} from discrete spaces to continuous spaces, which significantly extends the scope of parameter space symmetries for transformers. 

To further demonstrate the benefits of rotation symmetry for transformers, we explore its utility in \textbf{model fusion}. 
The goal of model fusion is to merge multiple well-trained end models in the parameter space to 
produce a single merged model with improved overall utility.
Model fusion is widely adopted across various settings, such as hyperparameter tuning (where end models are trained on the same benchmark) and multi-task learning (where end models are trained with different tasks)~\citep{jin2023dataless}.
Unlike ensemble learning~\citep{dietterich2000ensemble,lakshminarayanan2017simple,sagi2018ensemble,dong2020survey}, model fusion can work in a data-agnostic manner, 
making it suitable for privacy-sensitive scenarios such as federated learning~\citep{yurochkin2019bayesian,wang2020federated}.

The existing literature has demonstrated that the performance of model fusion is closely tied to the distance between the end models~\citep{wortsman2022model}.
Inspired by this finding, we propose a parameter matching algorithm that selects functionally equivalent end models from the equivalence class determined by rotation symmetry.
This approach ensures that the selected representative models are closer in parameter space, resulting in a smaller inner distance.
To achieve this, we formulate the problem of parameter matching as an optimization problem with orthogonal constraints.
Leveraging the continuous nature of rotation symmetry, we propose a closed-form solution to this problem. 
Our parameter matching algorithm is highly efficient, easy to implement, and can be seamlessly incorporated as a plug-and-play module for model fusion. 
To evaluate its effectiveness, we conduct extensive experiments with pre-trained transformers on real-world NLP and vision tasks. 
The experimental results demonstrate that incorporating rotation symmetry into parameter matching improves model fusion effectively and efficiently. 
Furthermore, additional experiments reveal that 
even matching a subset of parameters can lead to notable performance improvements, highlighting the practical utility of our approach. 
Our contributions are threefold:
\begin{itemize}[leftmargin=*]
\item We introduce a novel rotation symmetry for the attention mechanism in transformers, extending the concept of symmetry to a continuous space.
\item 
Building on rotation symmetry, we propose a theoretically optimal parameter matching algorithm that improves the effectiveness of model fusion in transformers.
\item 
Through extensive experiments, we validate the efficacy of our proposed parameter matching algorithm, demonstrating its potential to advance model fusion through parameter space symmetry.
\end{itemize}

\section{Preliminaries}
\nop{We first define some notations for further discussion.
Let $\bm{A}[i,:]$, $\bm{A}[:,j]$, and $\bm{A}[i,j]$ denote the $i$-th row vector, $j$-th column vector, and the $(i,j)$ element of the matrix $\bm{A}$, respectively.
Let $\bm{1}$ denote the all-ones vector.
We introduce the permutation symmetry of neural networks using MLP as an example. 
We first consider an $L$-layer MLP model}

To facilitate further discussion, we begin by defining the notations. 
Let $\bm{A}[i,:]$, $\bm{A}[:,j]$, and $\bm{A}[i,j]$ denote the $i$-th row vector, the $j$-th column vector, and the $(i,j)$-th element of the matrix $\bm{A}$, respectively. Additionally, let $\bm{1}$ denote an all-ones vector. We now illustrate the concept of permutation symmetry in neural networks, using an MLP as an example. Consider an $L$-layer MLP model defined as:
\begin{equation}\label{eq:mlp}
f_{\bm{W}}(\bm{X})=\bm{Z}^{(L)},\ \bm{Z}^{(l)}=\sigma(\bm{Z}^{(l-1)}(\bm{W}^{(l)})^\top+\bm{b}^{(l)}),
\end{equation}
where $\bm{Z}^{(0)}=\bm{X}$ is the input feature matrix, $\bm{b}^{(l)}$ is the bias vector ($\bm{b}^{(l)}$ is a row vector and can be seen as broadcast over all rows in \Cref{eq:mlp}), $\bm{W}=\{\bm{W}^{(l)},\bm{b}^{(l)}\}_{l=1,\dots,L}$ collects all learnable parameters, and $\sigma(\cdot)$ stands for a non-linear activation function, such as ReLU~\citep{nair2010rectified}. A loss function $\mathcal{L}(f_{\bm{W}}(\bm{X}),\bm{Y})$ is used to measure the distance between the model prediction and the ground truth label $\bm{Y}$. 

To analyze the permutation symmetry in the MLP model, let $\bm{P}\in\mathcal{P}$ be a permutation matrix, where $\bm{P}[i,j]\in\{0,1\}$ and $\bm{P}[i,:]\bm{1}=\bm{P}[:,j]^\top\bm{1}=1$ for any $i$, $j$.
All permutation matrices are orthogonal~\citep{strang1976linear}, satisfying $\bm{P}^\top = \bm{P}^{-1}$.
Consequently, for layer $l$ and $l+1$, we have:
\begin{equation}\label{eq:permutation_symmetry}
\scriptsize
\begin{aligned}
\bm{Z}^{(l+1)}&=\sigma\left(\sigma(\bm{Z}^{(l-1)}(\bm{W}^{(l)})^\top+\bm{b}^{(l)})(\bm{W}^{(l+1)})^\top+\bm{b}^{(l+1)}\right) \\
&=\sigma\left(\sigma(\bm{Z}^{(l-1)}(\bm{P}^\top\bm{W}^{(l)})^\top+\bm{b}^{(l)}\bm{P})(\bm{W}^{(l+1)}\bm{P})^\top+\bm{b}^{(l+1)}\right).
\end{aligned}
\end{equation}
The third equal sign holds because the element-wise activation function $\sigma$ is decoupled from column permutation (\textit{i.e.}, being multiplied by $\bm{P}$).
Based on \Cref{eq:permutation_symmetry}, it follows that for layers $l$ and $l+1$, the mappings
$\bm{W}^{(l)}\rightarrow\bm{P}^\top\bm{W}^{(l)}$, $\bm{b}^{(l)}\rightarrow\bm{b}^{(l)}\bm{P}$, $\bm{W}^{(l+1)}\rightarrow\bm{W}^{(l+1)}\bm{P}$ preserves the output $\bm{Z}^{(l+1)}$ for any input $\bm{Z}^{(l-1)}$.
For each pair of adjacent layers, a similar mapping exists independently based on a specific permutation matrix, denoted as $\bm{P}^{(l)}$ for layers $l$ and $l+1$. 
Consequently, for the entire $L$-layer MLP, there exists a mapping that preserves the model's predictions for any input $\bm{X}$:
$\bm{W}^{(l)}\rightarrow(\bm{P}^{(l)})^\top\bm{W}^{(l)}\bm{P}^{(l-1)}$, $\bm{b}^{(l)}\rightarrow\bm{b}^{(l)}\bm{P}^{(l)}$.
Equivalently, if we define
$\bm{W}^\prime=\{(\bm{P}^{(l)})^\top\bm{W}^{(l)}\bm{P}^{(l-1)},\bm{b}^{(l)}\bm{P}^{(l)}\}_{l=1,\dots,L}$ for any $\bm{P}^{(l)}\in\mathcal{P}$ ($l=1,\dots,L-1$) and $\bm{P}^{(0)}=\bm{P}^{(L)}=\bm{I}$, then we have $f_{\bm{W}^\prime}(\bm{X})=f_{\bm{W}}(\bm{X})$ for any $\bm{X}$.
This phenomenon is referred to as the \textit{permutation symmetry} of the parameter space~\citep{godfrey2022symmetries,hecht1990algebraic,navon2023equivariant,rossi2023permutation,simsek2021geometry}.
Leveraging permutation symmetry allows us to identify an equivalence class of functionally equivalent model parameters, which is known as \textit{permutation invariance}~\citep{ainsworth2023git,entezari2022role,lubana2023mechanistic}. 
We denote the equivalence relation induced by permutation invariance as $\pi$, where, for example, $\bm{W}^\prime = \pi(\bm{W})$ represents the equivalence between the original parameters $\bm{W}$ and the permuted parameters $\bm{W}^\prime$.

\section{Parameter Space Symmetry of Transformers}

\nop{Transformers~\citep{vaswani2017attention} have revolutionized the field of deep learning by offering a distinctive architecture that excels in handling sequential data, particularly in natural language processing (NLP) and beyond~\citep{brown2020language,devlin2019bert,liu2021swin,radford2021learning}. 
The power of transformers derives from its two main components\footnote{We follow the architecture in the original transformer paper~\citep{vaswani2017attention}.}, feedforward networks and self-attention layers~\citep{vaswani2017attention}.
To investigate the parameter space symmetry of transformers, we next focus on the two key modules separately.}

Transformers~\citep{vaswani2017attention} have revolutionized deep learning with their ability to handle sequential data effectively, particularly in natural language processing (NLP) and other fields~\citep{brown2020language,devlin2019bert,liu2021swin,radford2021learning}. Their success is primarily driven by two key components\footnote{We follow the architecture in the original transformer paper~\citep{vaswani2017attention}.}: feedforward networks and self-attention layers~\citep{vaswani2017attention}. To better understand the parameter space symmetry of transformers, we examine these two core modules individually in the following sections.

\subsection{Permutation Symmetry of Feedforward Networks}
We first look at the feedforward networks.
The feedforward network adopted in the transformer blocks is a two-layer MLP model which can be written as
\begin{equation}\label{eq:ffn}
\small
FFN(\bm{X})=LN(ReLU(\bm{X}\bm{W}_i^\top+\bm{b}_i)\bm{W}_o^\top+\bm{b}_o+\bm{X}),
\end{equation}
where $LN$ denotes the Layer Normalization operator~\citep{ba2016layer}.
Different from~\cite{vaswani2017attention}, we include the residual connection~\citep{he2016deep} and layer normalization modules into the formula of the feedforward network (and the self-attention layer mentioned later).
According to \Cref{eq:permutation_symmetry} and the analysis in Preliminaries, the feedforward networks have the permutation symmetry property and the equivalence class determined by permutation invariance is defined as
\begin{equation}\label{eq:match_ffn}
\bm{W}_i\rightarrow\bm{P}^\top\bm{W}_i,\ \bm{b}_i\rightarrow\bm{b}_i\bm{P},\ \bm{W}_o\rightarrow\bm{W}_o\bm{P},\ \bm{b}_o\rightarrow\bm{b}_o,
\end{equation}
where $\bm{P}\in\mathcal{P}$ is a permutation matrix. 
It is worth noting that the permutations of different feedforward networks in a transformer are independent due to the scalable modular design.
Consequently, we are able to flexibly compute the permutation invariance equivalence class of each $FFN$ module in a transformer model.

\subsection{Rotation Symmetry of Self-attention Layers}
We then focus on the self-attention layers.
In this paper, we introduce the \textit{rotation symmetry}
of self-attention layers.
For MLPs, we have to switch the order of multiplying $\bm{P}$ and passing the activation function $\sigma$ (the third equal sign in \Cref{eq:permutation_symmetry}), requiring the matrix $\bm{P}$ to be a permutation matrix.
In contrast, the self-attention layer does not contain an element-wise activation function, enabling a wider range of the matrix $\bm{P}$ in self-attention layers.
The self-attention layer can be written as
$$
\small
\begin{aligned}
ATTN(\bm{X})&=LN\left(Cat_{h=1}^H\left\{\bm{X}_{QKV}^h\right\}\bm{W}_O^\top+\bm{b}_O+\bm{X}\right), \\
\bm{X}_{QKV}^h&=Sftmx\left(\bm{X}_{Q}^h(\bm{X}_{K}^h)^\top/\sqrt{d_k}\right)\cdot\bm{X}_{V}^h,
\end{aligned}
$$
where $Cat$ stands for the operator concatenating the outputs of multi-head attention, $Sftmx(\cdot)$ denotes the softmax operator, $H$ denotes the number of multi-heads, and the subscripts $Q$, $K$, $V$, and $O$ denote Query, Key, Value, and Output, respectively.

\begin{figure}[t]
\centering
\includegraphics[clip, trim=3.5cm 5cm 3.5cm 5cm, width=0.9\linewidth]{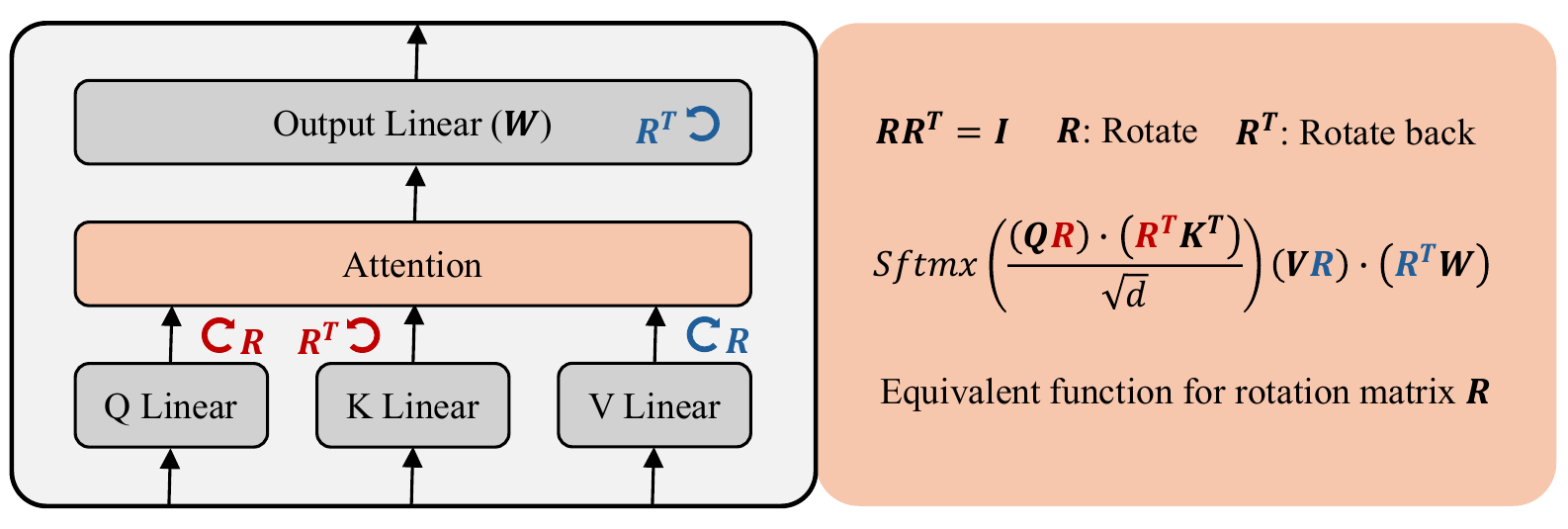}
\vspace{-3mm}
\caption{The rotation symmetry of self-attention layers.}
\label{fig:rotation_symmetry}
\vspace{-3mm}
\end{figure}

We first transform the query and key matrices as 
$$
\scriptsize
\begin{aligned}
\bm{X}_{Q}^h(\bm{X}_{K}^h)^\top&=(\bm{X}(\bm{W}_Q^h)^\top+\bm{b}_Q^h)(\bm{X}(\bm{W}_K^h)^\top+\bm{b}_K^h)^\top \\
&=(\bm{X}(\bm{W}_Q^h)^\top+\bm{b}_Q^h)\bm{R}\bm{R}^\top(\bm{X}(\bm{W}_K^h)^\top+\bm{b}_K^h)^\top \\
&=(\bm{X}(\bm{R}^\top\bm{W}_Q^h)^\top+\bm{b}_Q^h\bm{R})(\bm{X}(\bm{R}^\top\bm{W}_K^h)^\top+\bm{b}_K^h\bm{R})^\top,
\end{aligned}
$$
where $\bm{R}$ is a rotation matrix, \textit{i.e.}, $\bm{R}\bm{R}^\top=\bm{I}$. It is worth noting that each multi-head corresponds to a specific rotation matrix $\bm{R}$.
Let $\bm{W}_O=[\bm{W}_O^1\ \bm{W}_O^2\ \cdots\ \bm{W}_O^H]$ and we then rewrite the concatenating operation (with the product by $\bm{W}_O$) as $\sum_{h=1}^HS(\cdots)(\bm{X}(\bm{W}_V^h)^\top+\bm{b}_V^h)(\bm{W}_O^h)^\top$.
Similar to the Q-K case, we can transform the value and output matrices as $(\bm{X}(\bm{W}_V^h)^\top+\bm{b}_V^h)(\bm{W}_O^h)^\top=(\bm{X}(\bm{W}_V^h)^\top+\bm{b}_V^h)\bm{R}\bm{R}^\top(\bm{W}_O^h)^\top=(\bm{X}(\bm{R}^\top\bm{W}_V^h)^\top+\bm{b}_V^h\bm{R})(\bm{W}_O^h\bm{R})^\top$ for each multi-head $h$, where $\bm{R}$ is a rotation matrix.
Finally, we derive an equivalence class of the parameters in a self-attention layer determined by rotation invariance as
\begin{equation}\label{eq:match_attn}
\small
\begin{aligned}
&\bm{W}_Q^h\rightarrow(\bm{R}_{qk}^h)^\top\bm{W}_Q^h,\ \bm{b}_Q^h\rightarrow\bm{b}_Q^h\bm{R}_{qk}^h, \\ 
&\bm{W}_K^h\rightarrow(\bm{R}_{qk}^h)^\top\bm{W}_K^h,\ \bm{b}_K^h\rightarrow\bm{b}_K^h\bm{R}_{qk}^h, \bm{W}_V^h\rightarrow(\bm{R}_{vo}^h)^\top\bm{W}_V^h, \\ 
&\bm{b}_V^h\rightarrow\bm{b}_V^h\bm{R}_{vo}^h,\ \bm{W}_O^h\rightarrow\bm{W}_O^h\bm{R}_{vo}^h,\ \bm{b}_O\rightarrow\bm{b}_O,
\end{aligned}
\end{equation}
where $\bm{R}_{qk}^h$ and $\bm{R}_{vo}^h$ are rotation matrices for $h=1,\dots,H$.
The progress from permutation to rotation extends the symmetry of transformers to a continuous space and enhances our understanding of the parameter space symmetry of attention mechanism.
The denseness of the symmetry allows for the choice of better invariant models to analyze transformers' loss landscapes.
For better clarity, we provide an intuitive illustration of the rotation symmetry in \Cref{fig:rotation_symmetry}.

\section{Symmetry for Model Fusion}
In this section, we explore the benefit of the rotation symmetry of transformers in model fusion~\citep{li2023deep,matena2022merging,wortsman2022model,yadav2023ties,jin2023dataless,daheim2024model,yang2024adamerging}.
Model fusion is proposed to merge multiple given end models trained in different settings (\textit{e.g.}, upon different datasets and hyperparameter settings) in the parameter space to improve model utility and robustness.
Compared with ensemble learning~\citep{dietterich2000ensemble,sagi2018ensemble,lakshminarayanan2017simple}, model fusion has a lower inference-stage complexity without requiring access to the training data.
Most existing methods of model fusion conduct a weighted averaging of different end models, \textit{e.g.}, direct averaging~\citep{wortsman2022model}, Fisher-weighted averaging~\citep{matena2022merging}, and regression-mean averaging~\citep{jin2023dataless}.
We next show the potential of exploiting the permutation and rotation symmetry as a plug-and-play module to improve the model fusion techniques.

\subsection{Background and Motivation}
Let $\bm{W}_1,\dots,\bm{W}_k$ denote $k$ different end models (after training or pre-training) with the same architecture.
The goal of model fusion is to merge the given $k$ end models in the parameter space and obtain a single model.
If the given models are trained over different datasets, we can expect the merged model to have better utility and out-of-distribution robustness~\citep{jin2023dataless}.
The theoretical results in previous literature~\citep{wortsman2022model} have shown that \textit{strong convexity} and \textit{closer end models} can boost the utility of direct model fusion.
Consequently, the primary advantage of permutation (and rotation) symmetry applying to model fusion is that we can substitute the end models with corresponding carefully chosen equivalent models (in the equivalence class determined by permutation and rotation invariance) to make the selected end models more concentrated, \textit{i.e.}, closer to each other.
This step is usually called \textbf{parameter matching}~\citep{singh2020model,wang2020federated,ainsworth2023git}, aka parameter or neuron alignment.
Parameter matching, which aims to reduce the distance between end models, can naturally yield closer end models and improve model fusion performance.
On the other hand, different end models can lie in the basins of different local optimums regarding the highly non-convex nature of transformers.
\begin{figure}[t]
\centering
\includegraphics[width=0.8\linewidth]{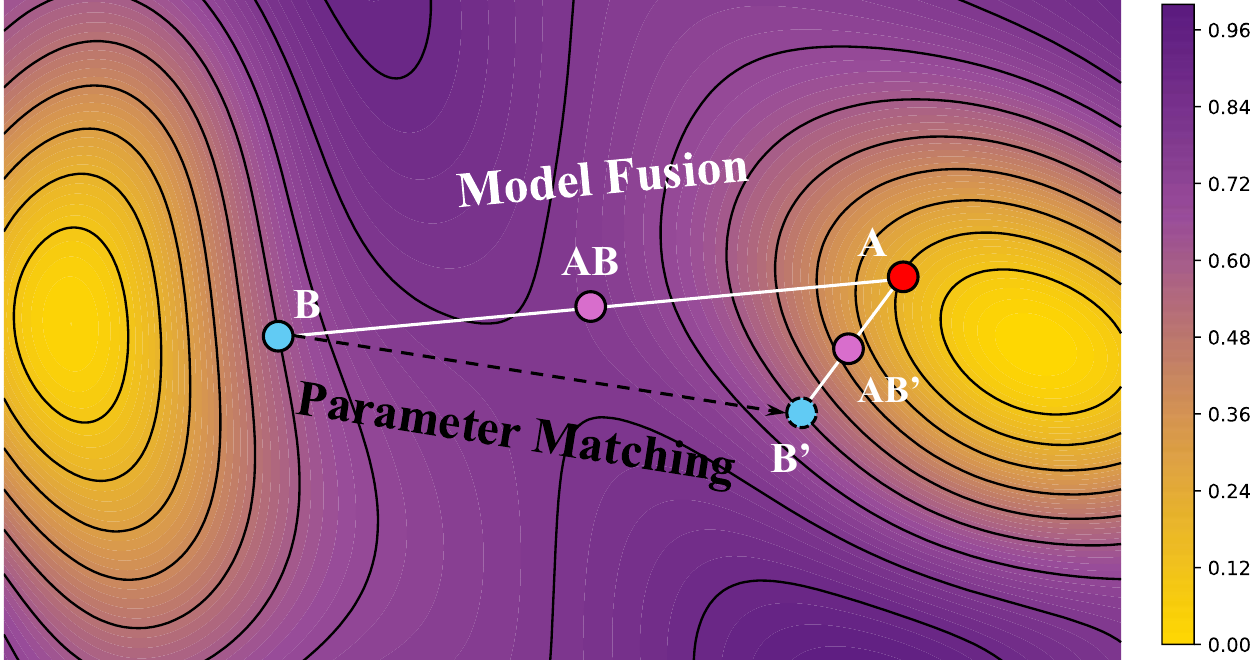}
\vspace{-3mm}
\caption{An intuitive example of the usage of parameter space symmetry for model fusion. The background shows the contour map of the loss landscape in the model parameter space. A and B are the original end models to be merged, AB is the result of naive model fusion, and AB' is the result of model fusion with parameter matching.}
\label{fig:model_fusion}
\vspace{-3mm}
\end{figure}
Previous studies have verified that parameter matching can merge different end models toward a single low-loss basin, \textit{i.e.}, the loss value along the linear interpolation between matched models shows an approximately flat or convex curve~\citep{entezari2022role,ainsworth2023git}. 
This property is called \textit{linear mode connectivity} and is regarded as a weak form of convexity~\citep{ainsworth2023git}.
As a result, parameter matching can also help improve the convexity of the objective in the area adjacent to the end models.
We showcase an intuitive example of using parameter space symmetry to improve model fusion in \Cref{fig:model_fusion} to illustrate the intuition behind parameter matching.
We can observe that the merged model after parameter matching (AB') has a lower loss value, \textit{i.e.}, better utility than the naive merged model (AB).

\subsection{Parameter Matching}
Next, our primary goal is to develop a practical parameter matching algorithm to minimize the distance between different end models. We begin by merging two models $\bm{W}_1$ and $\bm{W}_2$.

\paragraph{Matching two FFNs.}
Let $\{\bm{W}_{i_k},\bm{b}_{i_k},\bm{W}_{o_k},\bm{b}_{o_k}\}_{k=1,2}$ denote the model parameters in the two feedforward networks to be merged where $k$ denote the index of the networks.
We have already derived the equivalence relation of parameters in the feedforward networks determined by permutation invariance as \Cref{eq:match_ffn}. 
Based on \Cref{eq:match_ffn}, we can formulate parameter matching as an optimization problem as follows.
\begin{equation}\label{eq:obj_ffn}
\begin{aligned}
&\mathrm{argmin}_{\bm{P}_1,\bm{P}_2\in\mathcal{S}}\|\bm{P}_1^\top\bm{W}_{i_1}-\bm{P}_2^\top\bm{W}_{i_2}\|^2_F \\
&\quad+\|\bm{b}_{i_1}\bm{P}_1-\bm{b}_{i_2}\bm{P}_2\|^2_F+\|\bm{W}_{o_1}\bm{P}_1-\bm{W}_{o_2}\bm{P}_2\|^2_F.
\end{aligned}
\end{equation}
As shown in \Cref{eq:obj_ffn}, the goal is to find the functionally equivalent models from the equivalence classes $\pi(\bm{W}_1)$ and $\pi(\bm{W}_2)$ determined by $\bm{P}_1$ and $\bm{P}_2$, which has the smallest $\ell$-2 distance in the parameter space.
Following the method~\citep{ainsworth2023git}, we reformulate the optimization problem shown in \Cref{eq:obj_ffn} as a \textit{linear assignment problem}:
\begin{equation}\label{eq:obj_ffn_lap}
\mathrm{argmax}_{\bm{P}\in\mathcal{S}}\left<\bm{P},\bm{W}_{i_1}\bm{W}_{i_2}^\top+\bm{b}_{i_1}^\top\bm{b}_{i_2}+\bm{W}_{o_1}^\top\bm{W}_{o_2}\right>_F.
\end{equation}
Linear assignment problems such as~\Cref{eq:obj_ffn_lap} have been well-studied in previous literature~\citep{martello1987linear,burkard1999linear} and can be solved precisely by Hungarian Algorithm~\citep{martello1987linear}.
After calculating the value of $\bm{P}$, we substitute $\bm{P}=\bm{P}_1\bm{P}_2^{-1}$ back to \Cref{eq:match_ffn} and let $\bm{P}_2=\bm{I}$ (for simplicity), then we obtain the matched parameter as $\bm{W}_{i_1}\rightarrow\bm{P}^\top\bm{W}_{i_1},\ \bm{b}_{i_1}\rightarrow\bm{b}_{i_1}\bm{P},\ \bm{W}_{o_1}\rightarrow\bm{W}_{o_1}\bm{P},\ \bm{b}_{o_1}\rightarrow\bm{b}_{o_1}$.
Parameters of the second model $\{\bm{W}_{i_2},\bm{b}_{i_2},\bm{W}_{o_2},\bm{b}_{o_2}\}$ remain unchanged, which makes the second model an anchor model for matching.

\paragraph{Matching two ATTNs.}
Let the parameters in the two self-attention layers to be merged be $\{\bm{W}_{Q_k}^i,\bm{b}_{Q_k}^i,\bm{W}_{K_k}^i,\bm{b}_{K_k}^i,\bm{W}_{V_k}^i,\bm{b}_{V_k}^i,\bm{W}_{O_k}^i,\bm{b}_{O_k}^i\}_{k=1,2;i=1\dots H}$ where $k$ denote the index of the layers.
The equivalence relation of these parameters in the self-attention layers in terms of rotation invariance is shown in \Cref{eq:match_attn}.
Similar to matching FFNs, our goal is to match the parameters of one (source) attention layer to the other (target) attention layer where the matched parameters are supposed to be \textit{closest} to the target parameters while being functionally equivalent when feeding with any input data, \textit{i.e.}, in the equivalence class determined by rotation invariance. 
The goal can be formulated as an optimization problem. 
\begin{equation}\label{eq:obj_attn}
\small
\begin{aligned}
\min_{\bm{R}_{qk}^h,\bm{R}_{vo}^h\in\mathcal{R}}&\left\|\left[
\begin{array}{cc}
(\bm{W}_{Q_1}^h)^\top & (\bm{W}_{Q_2}^h)^\top \\
\bm{b}_{Q_1}^h & \bm{b}_{Q_2}^h \\
\end{array}\right]\left[
\begin{array}{c}
\bm{R}_{qk1}^h \\
-\bm{R}_{qk2}^h \\
\end{array}\right]\right\|_F^2 \\
+&\left\|\left[
\begin{array}{cc}
(\bm{W}_{K_1}^h)^\top & (\bm{W}_{K_2}^h)^\top \\
\bm{b}_{K_1}^h & \bm{b}_{K_2}^h \\
\end{array}\right]\left[
\begin{array}{c}
\bm{R}_{qk1}^h \\
-\bm{R}_{qk2}^h \\
\end{array}\right]\right\|_F^2 \\
+&\left\|\left[
\begin{array}{cc}
(\bm{W}_{V_1}^h)^\top & (\bm{W}_{V_2}^h)^\top \\
\bm{b}_{V_1}^h & \bm{b}_{V_2}^h \\
\end{array}\right]\left[
\begin{array}{c}
\bm{R}_{vo1}^h \\
-\bm{R}_{vo2}^h \\
\end{array}\right]\right\|_F^2 \\
+&\left\|\left[
\begin{array}{cc}
\bm{W}_{O_1}^h & \bm{W}_{O_2}^h \\
\bm{0} & \bm{0} \\
\end{array}\right]\left[
\begin{array}{c}
\bm{R}_{vo1}^h \\
-\bm{R}_{vo2}^h \\
\end{array}\right]\right\|_F^2,
\end{aligned}
\end{equation}
where $\mathcal{R}$ denotes the set of rotation matrices.
To solve this problem, we divide \Cref{eq:obj_attn} into two separate optimization problems (the first line in terms of $\bm{R}_{qk1}^h$, $\bm{R}_{qk2}^h$ as the first objective and the second line in terms of $\bm{R}_{vo1}^h$, $\bm{R}_{vo2}^h$ as the second objective).
Considering the similar formulation of these two optimization problems, in this paper, we propose the following theorem to solve both problems.
\begin{theorem}\label{thm:solution}
The following optimization problem has a closed-form solution.
\begin{equation}\label{eq:solution}
\begin{aligned}
\min_{\bm{R}_1,\bm{R}_2\in\mathcal{R}}&\left\|\left[
\begin{array}{cc}
\bm{W}_{Q_1}^\top & \bm{W}_{Q_2}^\top \\
\bm{b}_{Q_1} & \bm{b}_{Q_2} \\
\end{array}\right]\left[
\begin{array}{c}
\bm{R}_1 \\
-\bm{R}_2 \\
\end{array}\right]\right\|_F^2 \\
+&\left\|\left[
\begin{array}{cc}
\bm{W}_{K_1}^\top & \bm{W}_{K_2}^\top \\
\bm{b}_{K_1} & \bm{b}_{K_2} \\
\end{array}\right]\left[
\begin{array}{c}
\bm{R}_1 \\
-\bm{R}_2 \\
\end{array}\right]\right\|_F^2.
\end{aligned}
\end{equation}
The solution is given by
\begin{equation}
\bm{R}_1=\bm{U}\bm{V}^\top,\bm{R}_2=\bm{I},
\end{equation}
where $\bm{I}$ is the identity matrix and $\bm{U}\bm{\Sigma}\bm{V}^\top=\bm{W}_{Q_1}\bm{W}_{Q_2}^\top+\bm{W}_{K_1}\bm{W}_{K_2}^\top+\bm{b}_{Q_1}^\top\bm{b}_{Q_2}+\bm{b}_{K_1}^\top\bm{b}_{K_2}$ is the result of eigendecomposition.
\end{theorem}
We leave a detailed proof of \Cref{thm:solution} in \Cref{sec:appendix_proof}.
According to \Cref{thm:solution}, we can obtain the algorithm to match two self-attention layers shown in \Cref{alg:attn_matching}.
\Cref{alg:attn_matching} can be seen as an adaptation of the Kabsch algorithm~\citep{kabsch1976solution,umeyama1991least}.
Without loss of generality, we let the parameters in the second ($k=2$) self-attention layer be the anchor and conduct rotation for the first layer ($k=1$).
The denseness of rotation symmetry helps reduce the distance between the end models after parameter matching as \Cref{alg:attn_matching} shows.

\begin{algorithm}[t]
\caption{Matching two self-attention layers.}\label{alg:attn_matching}
\textbf{Input:} Model parameters of the attention layers \{$\bm{W}_{Q_k}^h$,$\bm{b}_{Q_k}^h$,$\bm{W}_{K_k}^h$,$\bm{b}_{K_k}^h$,$\bm{W}_{V_k}^h$,$\bm{b}_{V_k}^h$,$\bm{W}_{O_k}^h$,$\bm{b}_{O_k}^h$\}$_{k=1,2;h=1,\dots,H}$. \\
\textbf{Output:} The optimally matched parameters from source parameters ($k=1$) to anchor parameters ($k=2$, maintain unchanged).
\begin{algorithmic}[1]
\STATE QK solution: $\bm{R}_{qk1}^h=\bm{U}_{qk}^h(\bm{V}_{qk}^h)^\top,\bm{R}_{qk2}^h=\bm{I}$, where $\bm{U}_{qk}^h\bm{\Sigma}_{qk}^h(\bm{V}_{qk}^h)^\top=\bm{W}_{Q_1}^h(\bm{W}_{Q_2}^h)^\top+\bm{W}_{K_1}^h(\bm{W}_{K_2}^h)^\top+(\bm{b}_{Q_1}^h)^\top\bm{b}_{Q_2}^h+(\bm{b}_{K_1}^h)^\top\bm{b}_{K_2}^h$.
\STATE QK matching: $\bm{W}_{Q_1}^h\rightarrow(\bm{R}_{qk1}^h)^\top\bm{W}_{Q_1}^h$, $\bm{W}_{K_1}^h\rightarrow(\bm{R}_{qk1}^h)^\top\bm{W}_{K_1}^h$, $\bm{b}_{Q_1}^h\rightarrow\bm{b}_{Q_1}^h\bm{R}_{qk1}^h$, $\bm{b}_{K_1}^h\rightarrow\bm{b}_{K_1}^h\bm{R}_{qk1}^h$.
\STATE VO solution: $\bm{R}_{vo1}^h=\bm{U}_{vo}^h(\bm{V}_{vo}^h)^\top,\bm{R}_{vo2}^h=\bm{I}$, where $\bm{U}_{vo}^h\bm{\Sigma}_{vo}^h(\bm{V}_{vo}^h)^\top=\bm{W}_{V_1}^h(\bm{W}_{V_2}^h)^\top+(\bm{W}_{O_1}^h)^\top\bm{W}_{O_2}^h+(\bm{b}_{V_1}^h)^\top\bm{b}_{V_2}^h$.
\STATE VO matching: $\bm{W}_{V_1}^h\rightarrow(\bm{R}_{vo1}^h)^\top\bm{W}_{V_1}^h$, $\bm{W}_{O_1}^h\rightarrow\bm{W}_{O_1}^h\bm{R}_{vo1}^h$, $\bm{b}_{V_1}^h\rightarrow\bm{b}_{V_1}^h\bm{R}_{vo1}^h$.
\end{algorithmic}
\end{algorithm}

\paragraph{Rescaling Matching.}
We find that the rescaling symmetry~\citep{neyshabur2015path,meng2019g,godfrey2022symmetries,kalogeropoulos2024scale} can be integrated with our proposed rotation symmetry in the parameter matching algorithm.
For instance, for the (simplified) Q-K product $\bm{W}_Q\bm{W}_K$, we can find a rescaling operation that preserves the functionality of the Q-K product $a\bm{W}_Q\cdot\frac{1}{a}\bm{W}_K$ where $a\neq0$ is a real number. 
By adding a scalar variable to each parameter matrix, we can formulate the objective of rescaling symmetry as follows (taking the Q-K product as an example).
\begin{equation}\label{eq:rescaling_objective}
\begin{aligned}
\min_a&\left\|a\bm{W}_{Q_1}^\prime-\bm{W}_{Q_2}^\prime\right\|^2+\left\|a\bm{b}_{Q_1}^\prime-\bm{b}_{Q_2}^\prime\right\|^2 \\
&+\left\|\bm{W}_{K_1}^\prime/a-\bm{W}_{K_2}^\prime\right\|^2+\left\|\bm{b}_{K_1}^\prime/a-\bm{b}_{K_2}^\prime\right\|^2,
\end{aligned}
\end{equation}
where $a$ is the rescaling variable and $^\prime$ denotes the parameters after rotation symmetry-based matching.
Here, we still set model 2 as the anchor model and conduct the rescaling operation on model 1 to align with model 2.
We can easily solve the optimality condition of~\Cref{eq:rescaling_objective} as
\begin{equation}\label{eq:rescaling_derivative}
\begin{aligned}
\left(\left\|\bm{W}_{Q_1}^\prime\right\|^2+\left\|\bm{b}_{Q_1}^\prime\right\|^2\right)a^4-\left\|\bm{W}_{K_1}^\prime\right\|^2-\left\|\bm{b}_{K_1}^\prime\right\|^2& \\
-\left(\left<\bm{W}_{Q_1}^\prime,\bm{W}_{Q_2}^\prime\right>+\left<\bm{b}_{Q_1}^\prime,\bm{b}_{Q_2}^\prime\right>\right)a^3& \\
+\left(\left<\bm{W}_{K_1}^\prime,\bm{W}_{K_2}^\prime\right>+\left<\bm{b}_{K_1}^\prime,\bm{b}_{K_2}^\prime\right>\right)a&=0.
\end{aligned}
\end{equation}
The roots of \Cref{eq:rescaling_derivative} can be derived using numerical methods as the value of $a$.
We then conduct the rescaling operation to model 1 as $\bm{W}_{Q_1}^\prime\rightarrow a\bm{W}_{Q_1}^\prime$, $\bm{b}_{Q_1}^\prime\rightarrow a\bm{b}_{Q_1}^\prime$, $\bm{W}_{K_1}^\prime\rightarrow\frac{1}{a}\bm{W}_{K_1}^\prime$, $\bm{b}_{K_1}^\prime\rightarrow\frac{1}{a}\bm{b}_{K_1}^\prime$.
It is worth noting that the rescaling symmetry-based matching is conducted after the rotation symmetry-based matching.
Using the rescaling operation, we extend the rotation matrices to orthogonal matrices with larger norms.
Nevertheless, the end models are close to each other in practical scenarios so the value of $a$ is usually close to 1.

\paragraph{Complexity.}
We next provide a brief analysis of the complexity of our proposed parameter matching algorithm.
We let the hidden dimension of the target transformer be $d$, the layer of the target transformer be $L$, and the number of attention heads be $H$.
For the parameter matching of feedforward networks, the linear assignment problem can be solved in $O(d^3)$ by the Hungarian algorithm~\citep{kuhn1955hungarian,martello1987linear}.
Additionally, to solve the optimization problem in \Cref{eq:obj_attn}, \Cref{alg:attn_matching} requires the complexity of $O(d^3)$ for eigendecomposition. 
Hence, the complexity of our proposed full parameter matching algorithm of a transformer is $O(d^3LH)$, similar to the complexity of the feedforward.
It is worth noting that the complexity can be further reduced in two ways.
The first way is to match a subset of layers instead of all.
In this way, the complexity of our proposed parameter matching algorithm is $O(d^3L_sH)$ where $L_s$ denotes the number of selected layers.
The second way is to match each unit module (a single feedforward network or a single attention layer) in parallel.
The decoupling of matching different modules makes it easy to implement multiprocessing.
Consequently, the overall complexity becomes to $O(d^3LH/p)$ where $p$ is the number of processes in parallel.

\begin{table}[t]
\centering
\small
\tabcolsep = 1.6pt
\renewcommand{\arraystretch}{1.1}
\caption{Experimental results of in-domain (Emotion and NER) and out-of-domain (NER-CoNLL) model fusion for two base language models over Emotion and NER tasks.}
\label{tab:glue1}
\aboverulesep = 0pt
\belowrulesep = 0pt
\begin{tabular}{l|cc|cc|cc}
\toprule
 & \multicolumn{2}{c|}{Emotion} & \multicolumn{2}{c|}{NER} & \multicolumn{2}{c}{NER-CoNLL} \\
& Roberta & Deberta & Roberta & Deberta & Roberta & Deberta \\
\midrule
Simple & 35.87 & 2.99 & 60.88 & 27.54 & 26.86 & 10.80 \\
+match & 35.87 & 2.99 & 60.88 & \textbf{31.31} & \textbf{26.87} & \textbf{23.30} \\
\midrule
Fisher & 44.02 & 35.95 & 54.55 & 33.20 & \textbf{23.06} & 12.53 \\
+match & \textbf{44.05} & \textbf{35.98} & \textbf{54.58} & \textbf{33.83} & 23.05 & \textbf{18.33} \\
\midrule
Regmean & 35.87 & 2.99 & 60.88 & 27.54 & 26.86 & 10.80 \\
+match & \textbf{39.95} & 2.99 & 60.88 & \textbf{31.31} & \textbf{26.87} & \textbf{14.06} \\
\bottomrule
\end{tabular}
\vspace{-2mm}
\end{table}

\begin{table}[t]
\centering
\tabcolsep = 6pt
\caption{Experimental results of ViT merging over the image classification task.}
\label{tab:vit}
\aboverulesep = 0pt
\belowrulesep = 0pt
\begin{tabular}{l|ccc}
\toprule
 & w/o match & w/o rescale & match \\
\midrule
Simple & 7.60 & 10.22 & 10.19 \\
Fisher & 17.96 & 18.61 & 18.58 \\
Regmean & 14.24 & 15.31 & 15.35 \\
OT-ACTS-EMD & 32.08 & 32.50 & 32.53 \\
OT-ACTS & 61.15 & 61.23 & 61.25 \\
OT-WTS & 57.11 & 57.16 & 57.17 \\
\bottomrule
\end{tabular}
\end{table}


\begin{figure*}
    \centering
    \includegraphics[width=0.9\linewidth]{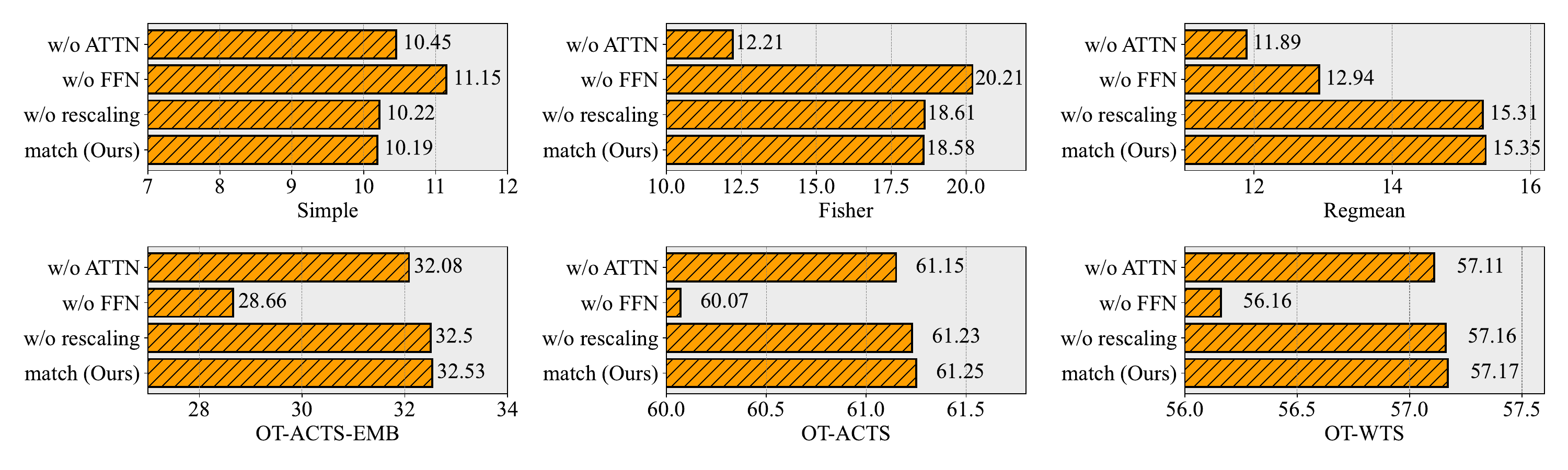}
    \vspace{-5mm}
    \caption{Ablation Study of ViT merging over the image classification task. ``ATTN'' is short for ``attention''. We compare our matching algorithm with its three variants (w/o ATTN/FFN/rescaling) and the original performance (w/o match) on all six merging baselines.}
    \label{fig:fig_sup_ablation}
    \vspace{-2mm}
\end{figure*}

\begin{figure}[t]
\centering
\includegraphics[width=0.9\linewidth]{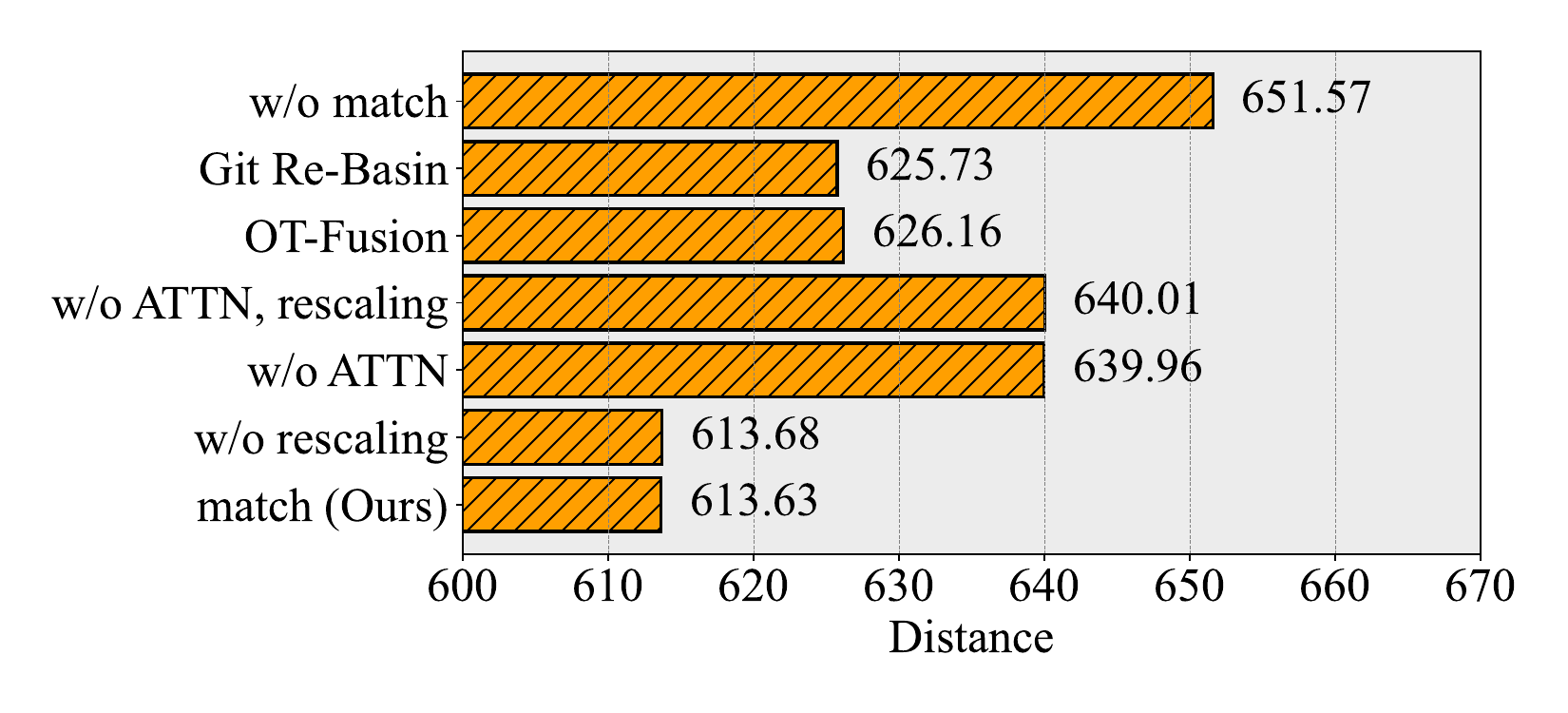}
\vspace{-5mm}
\caption{The Euclidean Distance of end ViT models after different parameter matching algorithms.}\label{tab:matching_distance}
\vspace{-2mm}
\end{figure}
\section{Experiments}

In this section, we conduct experiments to evaluate the effectiveness of our proposed parameter matching algorithm based on rotation symmetry. Specifically, we aim to answer the following research questions:
RQ1: Can our proposed parameter matching algorithm enhance the performance of model fusion for transformer-based models?
RQ2: How does rotation symmetry contribute to parameter matching for self-attention layers?
RQ3: How does parameter matching influence the loss landscape between independently trained models?
RQ4: As a plugin module, does our algorithm introduce significant additional computational overhead?
RQ5: Is matching all transformer layers equally important? Can we improve efficiency without compromising utility by matching only a subset of layers?

\subsection{Experimental Settings}\label{sec:exp_setting}
\paragraph{Platform.}
Our implementation is based on \texttt{Python 3.10} and \texttt{Pytorch 1.13}. All fine-tuning, parameter matching, and model merging processes are conducted on a cluster equipped with \texttt{Nvidia A100 80GB} GPUs.


\paragraph{Models.}

To evaluate the effectiveness of our approach, we use two widely adopted transformer models: RoBERTa~\citep{liu2019roberta} and DeBERTa~\citep{he2021deberta}.
We obtain pretrained models of RoBERTa (\texttt{roberta-base} with 12 attention layers) and DeBERTa-Large (\texttt{microsoft/deberta-v3-large} with 24 attention layers) from the Hugging Face library. 
For vision transformers (ViTs)~\citep{dosovitskiy2021imageworth16x16words}, we directly use the pretrained models following~\citep{imfeld2024transformer}. 
By selecting these three type of models, we assess the effectiveness of our method across different model scales and downstream tasks.


\paragraph{Finetuning and Matching.}

In the experiments, each pretrained language model is fine-tuned for 20 epochs on each dataset individually.
We set the learning rate at 1e-5, the batch size at 16, and the warmup ratio at 0.06 for each model. 
After fine-tuning, we perform parameter matching and model merging, considering both in-domain (pairwise fine-tuned models) and out-of-domain (grouped models) experiments. 
Notably, we match only the parameters within the attention layers, while the classifier module is directly copied from the model fine-tuned on the corresponding downstream task.
For additional details on datasets and baseline methods, please refer to~\Cref{app:b}.

\subsection{Performance of Model Fusion}\label{sec:performance}
To answer RQ1, we compare the performance of three model fusion baselines with and without our proposed parameter matching algorithm.




\paragraph{In-Domain Settings.}
We evaluate model fusion performance on Emotion Classification and Named Entity Recognition (NER) tasks, where our approach (\textbf{+match}) is integrated as a plugin module for three model fusion baselines.
For each task, we fine-tune the language models on each in-domain dataset (5 for Emotion, 6 for NER) respectively and merge them pairwise.
The first two columns in~\Cref{tab:glue1} present the average macro F1-score (for Emotion) and micro F1-score (for NER) of the merged models following~\citep{jin2023dataless}.

\paragraph{Out-of-Domain Settings.}
To assess the generalization ability, we merge models trained on all in-domain NER datasets and evaluate their performance on CoNLL datasets, which serve as out-of-domain (OOD) test sets.
The third column in~\Cref{tab:glue1} reports the OOD performance of the merged models.

\paragraph{ViT Settings.}
We employ three OT (optimal transportation)-based methods (\texttt{OT-ACTS-EMD}, \texttt{OT-ACTS}, and \texttt{OT-WTS}) from~\citep{imfeld2024transformer} as additional baselines.
Two pretrained models are merged, and we evaluate their performance on the CIFAR-10~\citep{krizhevsky2009learning} dataset.
\Cref{tab:vit} presents the classification accuracy of the merged models.
To improve the flexibility of parameter matching, we match each layer separately before conducting model fusion.
The merged model achieving the highest validation performance is selected as the final test model.

From the results in~\Cref{tab:glue1} and~\Cref{tab:vit}, we make the following observations:
(1) Our parameter matching algorithm consistently improves the performance of different model fusion methods. 
The reason is that parameter matching helps align distant models, bringing them closer in parameter space. 
As a result, the merged model is more likely to approach an overall minimum.
(2) Compared to RoBERTa-base, our method brings larger improvement for DeBERTa-large, suggesting that larger models benefit more from parameter matching. 
We explain this as larger models have more parameters that can be aligned, and in high-dimensional spaces, models tend to be more spread out. Subsequently, parameter matching helps bridge this gap more effectively.
(3) Among all the fusion methods, simple fusion shows the largest improvement after parameter matching. 
This is likely because simple fusion directly averages model weights, so reducing the distance between model parameters brings a clear benefit. 
In contrast, advanced fusion methods work by aligning outputs of the end models based on input data, which is equivalent to weighted averaging. Unlike direct averaging, these methods benefit from parameter matching in a more implicit way.

\subsection{Ablation Study}

To answer RQ2, we conduct ablation studies to evaluate the contributions of rotation symmetry in matching different model components. 
These experiments are performed on the same pair of pretrained ViT models used in the previous section.
We design three ablation settings:
(1) \underline{\textit{w/o ATTN}}: Matches only the FFN modules, excluding self-attention modules.
(2) \underline{\textit{w/o FFN}}: Matches only the self-attention modules, excluding FFN modules. 
(3) \underline{\textit{w/o rescaling}}: Disables the rescaling symmetry in our algorithm. 
Ablation results in \Cref{fig:fig_sup_ablation} demonstrate the following observations:
(1) In Fisher and RegMean, attention matching contributes more significantly to the fusion process. 
(2) In contrast, OTFusion benefits more from FFN matching. These distinctions demonstrate the nuanced interactions between the fusion mechanisms and corresponding component alignment strategies.
(3) \underline{\textit{w/o rescaling}} results in performance degradation across most settings, validating the effectiveness of rescaling symmetry integrated with rotation symmetry.

To further quantify the influence of rotation symmetry, we measure the Euclidean distance between the matched model and the anchor model in the parameter space. 
Instead of aggregating distances at the layer or module level, we analyze the entire model holistically to provide a more comprehensive assessment of alignment.
We also compare our method against two external baselines: Git Re-Basin~\citep{ainsworth2023git} and OT-Fusion~\citep{imfeld2024transformer}. 
Following the original paper, OT-Fusion is applied only to FFN modules.
The results in \Cref{tab:matching_distance} indicate that our rotation symmetry-based parameter matching algorithm consistently reduces the distance between end models more effectively than permutation symmetry-based methods.
This improvement can be attributed to the continuous nature of rotation symmetry, which allows for smoother and more precise alignment, especially in self-attention layers.
Furthermore, the integration of rescaling symmetry leads to additional distance reductions of end models, further enhancing parameter alignment.

\begin{figure}
    \centering
    \includegraphics[width=0.99\linewidth]{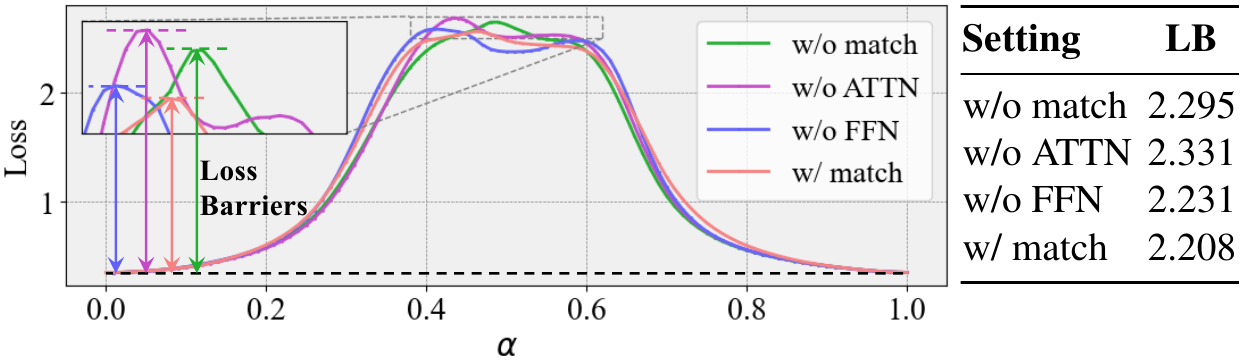}
    \vspace{-3mm}
    \caption{Loss landscapes and barriers between the two pretrained ViT models under four distinct matching settings. ``LB'' is short for ``Loss Barrier''.}
    \label{fig:loss_landscape}
    \vspace{-3mm}
\end{figure}

\begin{table}[t]
\centering
\tabcolsep = 5pt
\renewcommand{\arraystretch}{1.1}
\caption{The average runtime in seconds of the fine-tuning (top), matching (middle), and merging (bottom) stage. }
\label{tab:time}
\aboverulesep = 0pt
\belowrulesep = 0pt
\begin{tabular}{l|cccc}
\toprule
 & Deberta & Roberta & ViT  \\
\midrule
Fine-tuning & 12983.71 & 3071.49 & - \\
\hline
Matching & 1.59 & 1.71 & 3.48 \\
\hline
Simple Merging & 0.13 & 0.09 & 0.22  \\
Fisher Merging & 197.47 & 69.57 & 83.67  \\
Regmean Merging & 137.67 & 36.44 & 71.02  \\
\bottomrule
\end{tabular}
\vspace{-3mm}
\end{table}


\subsection{Loss Landscape Study}
To address RQ3, we examine the loss landscape between the two pretrained ViT models introduced in \Cref{sec:performance}, both with and without parameter matching. 
Specifically, we first analyze the cross entropy loss values $\mathcal{L}(\theta(\alpha))$ along the linear interpolation path between the end model parameters $\theta(\alpha) = \alpha \cdot \theta_A + (1 - \alpha) \cdot \theta_B, \quad \alpha \in [0, 1]$, where $\theta_A$ and $\theta_B$ denote the end models.
Next, we separately apply attention-only matching (\textit{w/o FFN}), FFN-only matching (\textit{w/o ATTN}), and complete matching (\textit{w/ match}) to $\theta_B$ before constructing the same linear interpolation path with respect to $\theta_A$. 
The results in \Cref{fig:loss_landscape} give rise to the following observations:
(1) All model pairs exhibit positive loss barriers, i.e., the range of loss value along the interpolation path, highlighting the strong inherent non-convexity of ViTs~\citep{park2022visiontransformerswork}.
(2) Complete matching yields the lowest loss barrier, indicating that aligning parameters through rotation symmetry leads to smoother connectivity between model pairs.
(3) While FFN-only matching (\textit{w/o ATTN}) has been proven effective for MLPs and CNNs~\citep{ainsworth2023git}, its effectiveness remains limited for Transformers. Their intricate loss landscape brings additional challenges for parameter matching. 

\subsection{Complexity Study}

To answer RQ4, we evaluate the computational overhead introduced by our parameter matching algorithm. 
Specifically, we measure the average runtime for fine-tuning (per dataset), matching (per model pair), and merging methods (per model pair) from the main experiments, as shown in \Cref{tab:time}.
For ViTs, we only use pre-trained models following~\cite{imfeld2024transformer} without fine-tuning.
The results show that our parameter matching module incurs an overhead of less than 5\% compared to Fisher / RegMean merging across all models, with negligible impact on overall complexity.

\begin{figure}[t]
\centering
\subfigure[Single-Layer Matching]{
\includegraphics[width=0.48\linewidth]{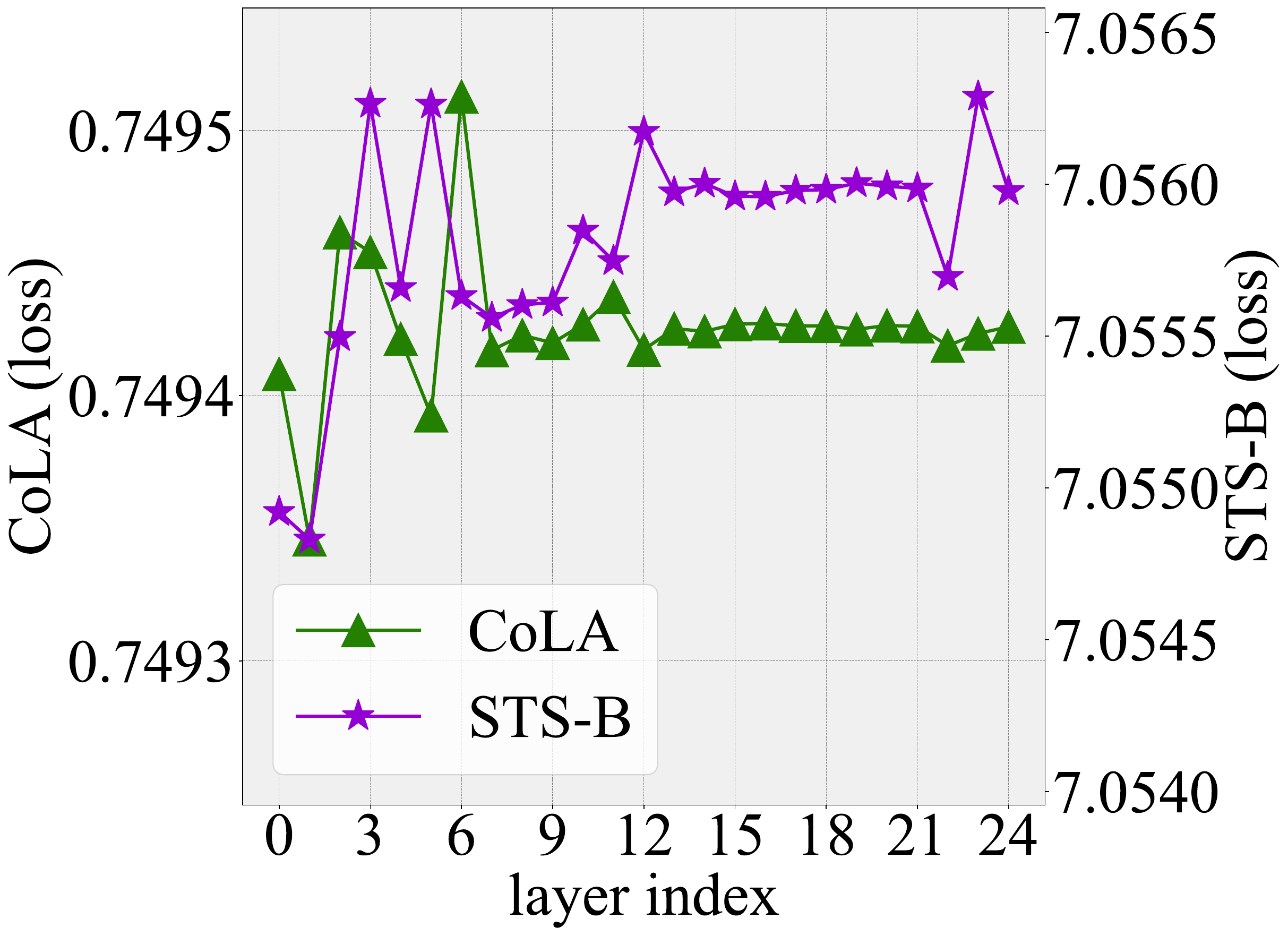}\label{fig:single}}
\subfigure[Tail-Layers Matching]{
\includegraphics[width=0.48\linewidth]{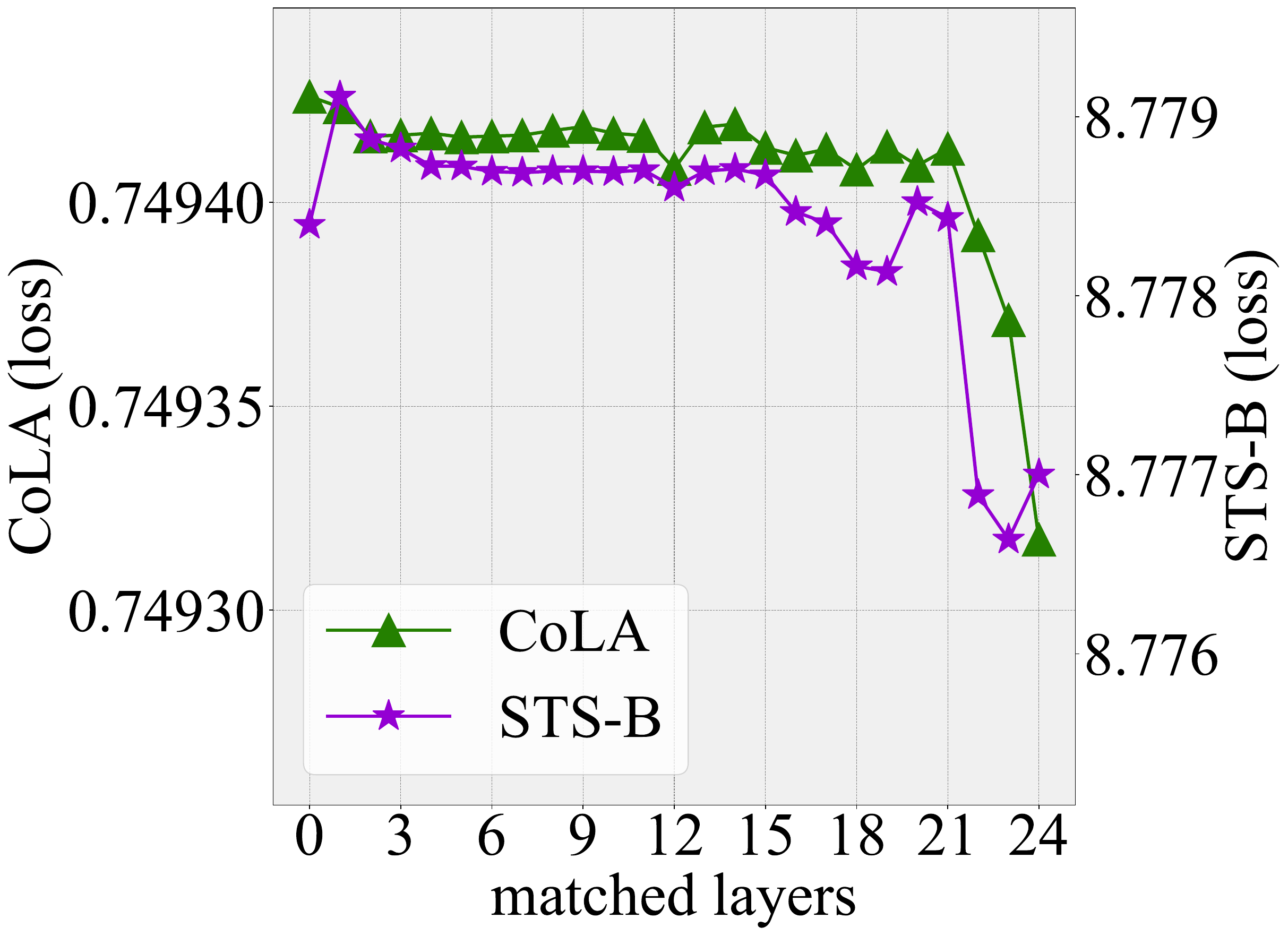}\label{fig:tail}}
\vspace{-3mm}
\caption{Evaluation loss of matching a subset of layers.}
\label{fig:subset_matching}
\vspace{-3mm}
\end{figure}

\subsection{Matching a Subset of Layers}




To answer RQ5, we investigate the effect of matching different subsets of layers. 
We fine-tune DeBERTa models on CoLA and STS-B from the GLUE benchmark and evaluate the performance of matched models under different subsets of layers. 
Specifically, we cross-validate the importance of each attention layer in matching through two settings:

\paragraph{Single-Layer Matching.}
In this setting, we match only a single attention layer while leaving all other layers unchanged. The evaluation loss corresponding to different matched layer indices is shown in \Cref{fig:single}.

\paragraph{Tail-Layers Matching.}
Here, we match a certain number of trailing attention layers (\textit{e.g.}, when matching three attention layers, we align only the last three layers). The evaluation loss for different numbers of matched layers is presented in \Cref{fig:tail}.

The experimental results in \Cref{fig:subset_matching} demonstrate that matching head layers yields greater improvements in model utility compared to tail layers. 
Notably, the loss value drops sharply when the first 5 layers are matched. 
Based on this observation, we can further improve efficiency by dropping tail layers without significantly compromising utility.

\section{Related Work}

\paragraph{Parameter Space Symmetry.}
\nop{The parameter space symmetry of deep neural networks has been studied for a long time.
Generally speaking, the parameter space symmetry is a set of models with different parameters but functionally equivalent. There are multiple ways to identify parameter space symmetries. Parameter space symmetries such as rescaling symmetry~\citep{neyshabur2015path,badrinarayanan2015symmetry,du2018algorithmic,meng2019g}, scaling symmetry~\citep{kunin2021neural}, and translation symmetry~\citep{kunin2021neural} had been identified in conventional deep neural networks to gain an in-depth understanding of the training dynamics and also accelerate the optimization process~\citep{zhao2022symmetry,zhao2023symmetries,zhao2024improving}. Another type of parameter space symmetry, \textit{i.e.}, permutation symmetry, was found to be closely related to the manifold of the global minimum and critical points~\citep{fukumizu2000local,brea2019weight,simsek2021geometry,benton2021loss,entezari2022role,ainsworth2023git}.}
Parameter space symmetry refers to a set of models with different parameter values but functionally equivalent. This concept has been extensively studied in the context of deep neural networks, as it plays a crucial role in understanding model behavior and training dynamics. 
Examples of parameter space symmetries include rescaling symmetry~\citep{neyshabur2015path,badrinarayanan2015symmetry,du2018algorithmic,meng2019g}, scaling symmetry~\citep{kunin2021neural}, and translation symmetry~\citep{kunin2021neural}. These symmetries have been identified in conventional deep neural networks to provide deeper insights into training dynamics and to accelerate the optimization process~\citep{zhao2022symmetry,zhao2023symmetries,zhao2024improving}. Another important type of parameter space symmetry is permutation symmetry, which has been shown to closely relate to the manifold of global minima and critical points~\citep{fukumizu2000local,brea2019weight,simsek2021geometry,benton2021loss,entezari2022role,ainsworth2023git}. The permutation symmetry can also be used to align (match) the outputs or model parameters of different end models with the same architecture~\citep{singh2020model,wang2020federated,ainsworth2023git,pena2023re,imfeld2024transformer,navon2024equivariant}.
Complementary to parameter space symmetry, recent advances in neural functionals and metanetworks~\citep{navon2023equivariant,zhou2023neural,zhou2023permutation,lim2024graph,tran2024equivariant} explore permutation-equivariant functionals that operate directly on model weights across diverse architectures.
Some concurrent works~\citep{ziyin2024symmetry,tran2024equivariant} investigate similar forms of rotation symmetry in neural networks.
\citet{ziyin2024symmetry} shows that the mirror symmetry leads to low-rankness. 
Meanwhile, \citet{tran2024equivariant} leverages the rotation symmetry to construct transformer-based neural functional networks.
In comparison, our study focuses on the role of rotation symmetry in model fusion and proposes a theoretically optimal parameter matching approach based on the properties of rotation symmetries.

\paragraph{Model Fusion.}
The goal of model fusion~\citep{li2023deep} is to merge multiple available end models (with the same architecture) to obtain a stronger model.
The scenarios of model fusion can be flexible.
When training on the same dataset, model fusion can be used to improve the model utility or generalization by merging models trained with different configurations or in different stages~\citep{izmailov2018averaging,gupta2020stochastic,cha2021swad,wortsman2022model,rame2022diverse,arpit2022ensemble,huang2024emr,yadav2024matters,hammoud2024model}.
As a representative method in this setting, ModelSoup~\citep{wortsman2022model} greedily averages the models fine-tuned with different hyperparameter configurations to improve the utility and robustness of the model.
In addition, when training on different datasets or tasks, model fusion can be used to improve out-of-domain generalization or multitasking of the model~\citep{matena2022merging,choshen2022fusing,li2022branch,jin2023dataless,zhou2024metagpt}, especially for language models.
A state-of-the-art merging algorithm, RegMean~\citep{jin2023dataless}, successfully merges language models fine-tuned over different tasks and improves the model's out-of-distribution generalization.
Moreover, model fusion plays a pivotal role in federated learning~\citep{konevcny2016federated,mcmahan2017communication,wang2020federated} when the local updates are collected to make a global update.
FedAvg~\citep{mcmahan2017communication} is a classical merging algorithm that directly computes the average of the local models as the updated global model.
Recent studies propose to incorporate the permutation symmetry to align the neurons of different end models~\citep{wang2020federated,singh2020model,ainsworth2023git}.
However, these methods fail to achieve a desirable performance when tackling transformer-based models~\citep{jin2023dataless}.

\section{Conclusion}
In this paper, we introduced rotation symmetry as a novel type of parameter space symmetry for transformers, extending the concept of permutation symmetry to continuous spaces. 
Building on this foundation, we proposed a theoretically optimal parameter matching algorithm to enhance the fusion of transformer models in a plug-and-play manner.
To validate our approach, we conducted extensive experiments on real-world NLP and vision benchmarks. 
The results demonstrated that incorporating rotation symmetry effectively and efficiently facilitates transformer model fusion, showcasing our method's practical utility.

\section*{Impact Statement}
Our study provides novel insights into the parameter space symmetry of transformers and establishes a practical framework for advancing model fusion techniques. 
Bridging theoretical innovations with practical applications, this work reveals the potential for leveraging parameter space symmetries in deep learning research.
There are many potential societal consequences of our work, none of which we feel must be specifically highlighted here.

\section*{Acknowledgment}
This work is supported in part by the National Science Foundation (NSF) under grants IIS-2006844, IIS-2144209, IIS-2223769, CNS-2154962, BCS-2228534, and CMMI-2411248; the Office of Naval Research (ONR) under grant N000142412636; and the Commonwealth Cyber Initiative (CCI) under grant VV-1Q24-011.

\bibliography{icml2025}

\begin{thebibliography}{100}
\providecommand{\natexlab}[1]{#1}
\providecommand{\url}[1]{\texttt{#1}}
\expandafter\ifx\csname urlstyle\endcsname\relax
  \providecommand{\doi}[1]{doi: #1}\else
  \providecommand{\doi}{doi: \begingroup \urlstyle{rm}\Url}\fi

\bibitem[Ainsworth et~al.(2023)Ainsworth, Hayase, and Srinivasa]{ainsworth2023git}
Ainsworth, S., Hayase, J., and Srinivasa, S.
\newblock Git re-basin: Merging models modulo permutation symmetries.
\newblock In \emph{International Conference on Learning Representations}, 2023.

\bibitem[Armenta \& Jodoin(2021)Armenta and Jodoin]{armenta2021representation}
Armenta, M. and Jodoin, P.-M.
\newblock The representation theory of neural networks.
\newblock \emph{Mathematics}, 9\penalty0 (24):\penalty0 3216, 2021.

\bibitem[Arpit et~al.(2022)Arpit, Wang, Zhou, and Xiong]{arpit2022ensemble}
Arpit, D., Wang, H., Zhou, Y., and Xiong, C.
\newblock Ensemble of averages: Improving model selection and boosting performance in domain generalization.
\newblock \emph{Advances in Neural Information Processing Systems}, 35:\penalty0 8265--8277, 2022.

\bibitem[Ba et~al.(2016)Ba, Kiros, and Hinton]{ba2016layer}
Ba, J.~L., Kiros, J.~R., and Hinton, G.~E.
\newblock Layer normalization.
\newblock \emph{arXiv preprint arXiv:1607.06450}, 2016.

\bibitem[Badrinarayanan et~al.(2015)Badrinarayanan, Mishra, and Cipolla]{badrinarayanan2015symmetry}
Badrinarayanan, V., Mishra, B., and Cipolla, R.
\newblock Symmetry-invariant optimization in deep networks.
\newblock \emph{arXiv preprint arXiv:1511.01754}, 2015.

\bibitem[Benton et~al.(2021)Benton, Maddox, Lotfi, and Wilson]{benton2021loss}
Benton, G., Maddox, W., Lotfi, S., and Wilson, A. G.~G.
\newblock Loss surface simplexes for mode connecting volumes and fast ensembling.
\newblock In \emph{International Conference on Machine Learning}, pp.\  769--779, 2021.

\bibitem[Bostan \& Klinger(2018)Bostan and Klinger]{bostan-klinger-2018-analysis}
Bostan, L.-A.-M. and Klinger, R.
\newblock An analysis of annotated corpora for emotion classification in text.
\newblock In Bender, E.~M., Derczynski, L., and Isabelle, P. (eds.), \emph{Proceedings of the 27th International Conference on Computational Linguistics}, pp.\  2104--2119, Santa Fe, New Mexico, USA, August 2018. Association for Computational Linguistics.
\newblock URL \url{https://aclanthology.org/C18-1179}.

\bibitem[Brea et~al.(2019)Brea, Simsek, Illing, and Gerstner]{brea2019weight}
Brea, J., Simsek, B., Illing, B., and Gerstner, W.
\newblock Weight-space symmetry in deep networks gives rise to permutation saddles, connected by equal-loss valleys across the loss landscape.
\newblock \emph{arXiv preprint arXiv:1907.02911}, 2019.

\bibitem[Brown et~al.(2020)Brown, Mann, Ryder, Subbiah, Kaplan, Dhariwal, Neelakantan, Shyam, Sastry, Askell, et~al.]{brown2020language}
Brown, T., Mann, B., Ryder, N., Subbiah, M., Kaplan, J.~D., Dhariwal, P., Neelakantan, A., Shyam, P., Sastry, G., Askell, A., et~al.
\newblock Language models are few-shot learners.
\newblock \emph{Advances in neural information processing systems}, 33:\penalty0 1877--1901, 2020.

\bibitem[Burkard \& Cela(1999)Burkard and Cela]{burkard1999linear}
Burkard, R.~E. and Cela, E.
\newblock Linear assignment problems and extensions.
\newblock In \emph{Handbook of combinatorial optimization: Supplement volume A}, pp.\  75--149. 1999.

\bibitem[Cha et~al.(2021)Cha, Chun, Lee, Cho, Park, Lee, and Park]{cha2021swad}
Cha, J., Chun, S., Lee, K., Cho, H.-C., Park, S., Lee, Y., and Park, S.
\newblock Swad: Domain generalization by seeking flat minima.
\newblock \emph{Advances in Neural Information Processing Systems}, 34:\penalty0 22405--22418, 2021.

\bibitem[Choshen et~al.(2022)Choshen, Venezian, Slonim, and Katz]{choshen2022fusing}
Choshen, L., Venezian, E., Slonim, N., and Katz, Y.
\newblock Fusing finetuned models for better pretraining.
\newblock \emph{arXiv preprint arXiv:2204.03044}, 2022.

\bibitem[Clark et~al.(2020)Clark, Luong, Le, and Manning]{clark2020electra}
Clark, K., Luong, M.-T., Le, Q.~V., and Manning, C.~D.
\newblock Electra: Pre-training text encoders as discriminators rather than generators.
\newblock In \emph{International Conference on Learning Representations}, 2020.

\bibitem[Daheim et~al.(2024)Daheim, M{\"o}llenhoff, Ponti, Gurevych, and Khan]{daheim2024model}
Daheim, N., M{\"o}llenhoff, T., Ponti, E., Gurevych, I., and Khan, M.~E.
\newblock Model merging by uncertainty-based gradient matching.
\newblock In \emph{International Conference on Learning Representations}, 2024.

\bibitem[Devlin et~al.(2019)Devlin, Chang, Lee, and Toutanova]{devlin2019bert}
Devlin, J., Chang, M.-W., Lee, K., and Toutanova, K.
\newblock Bert: Pre-training of deep bidirectional transformers for language understanding.
\newblock In \emph{Proceedings of the 2019 Conference of the North American Chapter of the Association for Computational Linguistics: Human Language Technologies, Volume 1 (Long and Short Papers)}, pp.\  4171--4186, 2019.

\bibitem[Dietterich(2000)]{dietterich2000ensemble}
Dietterich, T.~G.
\newblock Ensemble methods in machine learning.
\newblock In \emph{International workshop on multiple classifier systems}, pp.\  1--15, 2000.

\bibitem[Dong et~al.(2020)Dong, Yu, Cao, Shi, and Ma]{dong2020survey}
Dong, X., Yu, Z., Cao, W., Shi, Y., and Ma, Q.
\newblock A survey on ensemble learning.
\newblock \emph{Frontiers of Computer Science}, 14:\penalty0 241--258, 2020.

\bibitem[Dosovitskiy et~al.(2021)Dosovitskiy, Beyer, Kolesnikov, Weissenborn, Zhai, Unterthiner, Dehghani, Minderer, Heigold, Gelly, Uszkoreit, and Houlsby]{dosovitskiy2021imageworth16x16words}
Dosovitskiy, A., Beyer, L., Kolesnikov, A., Weissenborn, D., Zhai, X., Unterthiner, T., Dehghani, M., Minderer, M., Heigold, G., Gelly, S., Uszkoreit, J., and Houlsby, N.
\newblock An image is worth 16x16 words: Transformers for image recognition at scale, 2021.
\newblock URL \url{https://arxiv.org/abs/2010.11929}.

\bibitem[Du et~al.(2018)Du, Hu, and Lee]{du2018algorithmic}
Du, S.~S., Hu, W., and Lee, J.~D.
\newblock Algorithmic regularization in learning deep homogeneous models: Layers are automatically balanced.
\newblock \emph{Advances in neural information processing systems}, 31, 2018.

\bibitem[Entezari et~al.(2022)Entezari, Sedghi, Saukh, and Neyshabur]{entezari2022role}
Entezari, R., Sedghi, H., Saukh, O., and Neyshabur, B.
\newblock The role of permutation invariance in linear mode connectivity of neural networks.
\newblock In \emph{International Conference on Learning Representations}, 2022.

\bibitem[Ferbach et~al.(2024)Ferbach, Goujaud, Gidel, and Dieuleveut]{ferbach2024proving}
Ferbach, D., Goujaud, B., Gidel, G., and Dieuleveut, A.
\newblock Proving linear mode connectivity of neural networks via optimal transport.
\newblock In \emph{International Conference on Artificial Intelligence and Statistics}, pp.\  3853--3861, 2024.

\bibitem[Fukumizu \& Amari(2000)Fukumizu and Amari]{fukumizu2000local}
Fukumizu, K. and Amari, S.-i.
\newblock Local minima and plateaus in hierarchical structures of multilayer perceptrons.
\newblock \emph{Neural networks}, 13\penalty0 (3):\penalty0 317--327, 2000.

\bibitem[Glorot et~al.(2011)Glorot, Bordes, and Bengio]{glorot2011deep}
Glorot, X., Bordes, A., and Bengio, Y.
\newblock Deep sparse rectifier neural networks.
\newblock In \emph{Proceedings of the fourteenth international conference on artificial intelligence and statistics}, pp.\  315--323, 2011.

\bibitem[Godfrey et~al.(2022)Godfrey, Brown, Emerson, and Kvinge]{godfrey2022symmetries}
Godfrey, C., Brown, D., Emerson, T., and Kvinge, H.
\newblock On the symmetries of deep learning models and their internal representations.
\newblock \emph{Advances in Neural Information Processing Systems}, 35:\penalty0 11893--11905, 2022.

\bibitem[Gower \& Dijksterhuis(2004)Gower and Dijksterhuis]{gower2004procrustes}
Gower, J.~C. and Dijksterhuis, G.~B.
\newblock \emph{Procrustes problems}, volume~30.
\newblock OUP Oxford, 2004.

\bibitem[Grigsby et~al.(2023)Grigsby, Lindsey, and Rolnick]{grigsby2023hidden}
Grigsby, E., Lindsey, K., and Rolnick, D.
\newblock Hidden symmetries of relu networks.
\newblock In \emph{International Conference on Machine Learning}, pp.\  11734--11760, 2023.

\bibitem[Gupta et~al.(2020)Gupta, Serrano, and DeCoste]{gupta2020stochastic}
Gupta, V., Serrano, S.~A., and DeCoste, D.
\newblock Stochastic weight averaging in parallel: Large-batch training that generalizes well.
\newblock In \emph{International Conference on Learning Representations}, 2020.

\bibitem[Hammoud et~al.(2024)Hammoud, Michieli, Pizzati, Torr, Bibi, Ghanem, and Ozay]{hammoud2024model}
Hammoud, H., Michieli, U., Pizzati, F., Torr, P., Bibi, A., Ghanem, B., and Ozay, M.
\newblock Model merging and safety alignment: One bad model spoils the bunch.
\newblock In \emph{Findings of the Association for Computational Linguistics: EMNLP 2024}, pp.\  13033--13046, 2024.

\bibitem[He et~al.(2016)He, Zhang, Ren, and Sun]{he2016deep}
He, K., Zhang, X., Ren, S., and Sun, J.
\newblock Deep residual learning for image recognition.
\newblock In \emph{Proceedings of the IEEE conference on computer vision and pattern recognition}, pp.\  770--778, 2016.

\bibitem[He et~al.(2021)He, Liu, Gao, and Chen]{he2021deberta}
He, P., Liu, X., Gao, J., and Chen, W.
\newblock Deberta: Decoding-enhanced bert with disentangled attention, 2021.

\bibitem[He et~al.(2024)He, Zheng, Soga, Zhu, Dong, and Li]{he2024explaining}
He, Y., Zheng, Z., Soga, P., Zhu, Y., Dong, Y., and Li, J.
\newblock Explaining graph neural networks with large language models: A counterfactual perspective on molecule graphs.
\newblock In \emph{Findings of the Association for Computational Linguistics: EMNLP 2024}, pp.\  7079--7096, 2024.

\bibitem[Hecht-Nielsen(1990)]{hecht1990algebraic}
Hecht-Nielsen, R.
\newblock On the algebraic structure of feedforward network weight spaces.
\newblock In \emph{Advanced Neural Computers}, pp.\  129--135. 1990.

\bibitem[Hovy et~al.(2006)Hovy, Marcus, Palmer, Ramshaw, and Weischedel]{hovy-etal-2006-ontonotes}
Hovy, E., Marcus, M., Palmer, M., Ramshaw, L., and Weischedel, R.
\newblock {O}nto{N}otes: The 90{\%} solution.
\newblock In Moore, R.~C., Bilmes, J., Chu-Carroll, J., and Sanderson, M. (eds.), \emph{Proceedings of the Human Language Technology Conference of the {NAACL}, Companion Volume: Short Papers}, pp.\  57--60, New York City, USA, June 2006. Association for Computational Linguistics.
\newblock URL \url{https://aclanthology.org/N06-2015}.

\bibitem[Huang et~al.(2024)Huang, Ye, Chen, He, Yue, and Ouyang]{huang2024emr}
Huang, C., Ye, P., Chen, T., He, T., Yue, X., and Ouyang, W.
\newblock Emr-merging: Tuning-free high-performance model merging.
\newblock \emph{Advances in Neural Information Processing Systems}, 37:\penalty0 122741--122769, 2024.

\bibitem[Imfeld et~al.(2024)Imfeld, Graldi, Giordano, Hofmann, Anagnostidis, and Singh]{imfeld2024transformer}
Imfeld, M., Graldi, J., Giordano, M., Hofmann, T., Anagnostidis, S., and Singh, S.~P.
\newblock Transformer fusion with optimal transport.
\newblock In \emph{International Conference on Learning Representations}, 2024.

\bibitem[Izmailov et~al.(2018)Izmailov, Podoprikhin, Garipov, Vetrov, and Wilson]{izmailov2018averaging}
Izmailov, P., Podoprikhin, D., Garipov, T., Vetrov, D., and Wilson, A.~G.
\newblock Averaging weights leads to wider optima and better generalization.
\newblock In \emph{34th Conference on Uncertainty in Artificial Intelligence (UAI)}, pp.\  876--885, 2018.

\bibitem[Jin et~al.(2023)Jin, Ren, Preotiuc-Pietro, and Cheng]{jin2023dataless}
Jin, X., Ren, X., Preotiuc-Pietro, D., and Cheng, P.
\newblock Dataless knowledge fusion by merging weights of language models.
\newblock In \emph{International Conference on Learning Representations}, 2023.

\bibitem[Kabsch(1976)]{kabsch1976solution}
Kabsch, W.
\newblock A solution for the best rotation to relate two sets of vectors.
\newblock \emph{Acta Crystallographica Section A: Crystal Physics, Diffraction, Theoretical and General Crystallography}, 32\penalty0 (5):\penalty0 922--923, 1976.

\bibitem[Kalogeropoulos et~al.(2024)Kalogeropoulos, Bouritsas, and Panagakis]{kalogeropoulos2024scale}
Kalogeropoulos, I., Bouritsas, G., and Panagakis, Y.
\newblock Scale equivariant graph metanetworks.
\newblock \emph{Advances in neural information processing systems}, 37:\penalty0 106800--106840, 2024.

\bibitem[Kone{\v{c}}n{\`y} et~al.(2016)Kone{\v{c}}n{\`y}, McMahan, Yu, Richt{\'a}rik, Suresh, and Bacon]{konevcny2016federated}
Kone{\v{c}}n{\`y}, J., McMahan, H.~B., Yu, F.~X., Richt{\'a}rik, P., Suresh, A.~T., and Bacon, D.
\newblock Federated learning: Strategies for improving communication efficiency.
\newblock \emph{arXiv preprint arXiv:1610.05492}, 2016.

\bibitem[Krizhevsky et~al.(2009)Krizhevsky, Hinton, et~al.]{krizhevsky2009learning}
Krizhevsky, A., Hinton, G., et~al.
\newblock Learning multiple layers of features from tiny images.
\newblock 2009.

\bibitem[Kuhn(1955)]{kuhn1955hungarian}
Kuhn, H.~W.
\newblock The hungarian method for the assignment problem.
\newblock \emph{Naval research logistics quarterly}, 2\penalty0 (1-2):\penalty0 83--97, 1955.

\bibitem[Kunin et~al.(2021)Kunin, Sagastuy-Brena, Ganguli, Yamins, and Tanaka]{kunin2021neural}
Kunin, D., Sagastuy-Brena, J., Ganguli, S., Yamins, D.~L., and Tanaka, H.
\newblock Neural mechanics: Symmetry and broken conservation laws in deep learning dynamics.
\newblock In \emph{International Conference on Learning Representations}, 2021.

\bibitem[Lakshminarayanan et~al.(2017)Lakshminarayanan, Pritzel, and Blundell]{lakshminarayanan2017simple}
Lakshminarayanan, B., Pritzel, A., and Blundell, C.
\newblock Simple and scalable predictive uncertainty estimation using deep ensembles.
\newblock \emph{Advances in neural information processing systems}, 30, 2017.

\bibitem[Lewis et~al.(2020)Lewis, Liu, Goyal, Ghazvininejad, Mohamed, Levy, Stoyanov, and Zettlemoyer]{lewis2020bart}
Lewis, M., Liu, Y., Goyal, N., Ghazvininejad, M., Mohamed, A., Levy, O., Stoyanov, V., and Zettlemoyer, L.
\newblock Bart: Denoising sequence-to-sequence pre-training for natural language generation, translation, and comprehension.
\newblock In \emph{Proceedings of the 58th Annual Meeting of the Association for Computational Linguistics}, pp.\  7871--7880, 2020.

\bibitem[Li et~al.(2022)Li, Gururangan, Dettmers, Lewis, Althoff, Smith, and Zettlemoyer]{li2022branch}
Li, M., Gururangan, S., Dettmers, T., Lewis, M., Althoff, T., Smith, N.~A., and Zettlemoyer, L.
\newblock Branch-train-merge: Embarrassingly parallel training of expert language models.
\newblock In \emph{First Workshop on Interpolation Regularizers and Beyond at NeurIPS 2022}, 2022.

\bibitem[Li et~al.(2023)Li, Peng, Zhang, Ding, Hu, and Shen]{li2023deep}
Li, W., Peng, Y., Zhang, M., Ding, L., Hu, H., and Shen, L.
\newblock Deep model fusion: A survey.
\newblock \emph{arXiv preprint arXiv:2309.15698}, 2023.

\bibitem[Lim et~al.(2024{\natexlab{a}})Lim, Maron, Law, Lorraine, and Lucas]{lim2024graph}
Lim, D., Maron, H., Law, M.~T., Lorraine, J., and Lucas, J.
\newblock Graph metanetworks for processing diverse neural architectures.
\newblock In \emph{International Conference on Learning Representations}, 2024{\natexlab{a}}.

\bibitem[Lim et~al.(2024{\natexlab{b}})Lim, Putterman, Walters, Maron, and Jegelka]{lim2024empirical}
Lim, D., Putterman, T., Walters, R., Maron, H., and Jegelka, S.
\newblock The empirical impact of neural parameter symmetries, or lack thereof.
\newblock \emph{Advances in Neural Information Processing Systems}, 37:\penalty0 28322--28358, 2024{\natexlab{b}}.

\bibitem[Liu et~al.(2019)Liu, Ott, Goyal, Du, Joshi, Chen, Levy, Lewis, Zettlemoyer, and Stoyanov]{liu2019roberta}
Liu, Y., Ott, M., Goyal, N., Du, J., Joshi, M., Chen, D., Levy, O., Lewis, M., Zettlemoyer, L., and Stoyanov, V.
\newblock Roberta: A robustly optimized bert pretraining approach.
\newblock \emph{arXiv preprint arXiv:1907.11692}, 2019.

\bibitem[Liu(2024)]{ziyin2024symmetry}
Liu, Z.
\newblock Symmetry induces structure and constraint of learning.
\newblock In \emph{International Conference on Machine Learning}, 2024.

\bibitem[Liu et~al.(2021)Liu, Lin, Cao, Hu, Wei, Zhang, Lin, and Guo]{liu2021swin}
Liu, Z., Lin, Y., Cao, Y., Hu, H., Wei, Y., Zhang, Z., Lin, S., and Guo, B.
\newblock Swin transformer: Hierarchical vision transformer using shifted windows.
\newblock In \emph{Proceedings of the IEEE/CVF international conference on computer vision}, pp.\  10012--10022, 2021.

\bibitem[Lubana et~al.(2023)Lubana, Bigelow, Dick, Krueger, and Tanaka]{lubana2023mechanistic}
Lubana, E.~S., Bigelow, E.~J., Dick, R.~P., Krueger, D., and Tanaka, H.
\newblock Mechanistic mode connectivity.
\newblock In \emph{International Conference on Machine Learning}, pp.\  22965--23004, 2023.

\bibitem[Martello \& Toth(1987)Martello and Toth]{martello1987linear}
Martello, S. and Toth, P.
\newblock Linear assignment problems.
\newblock In \emph{North-Holland Mathematics Studies}, volume 132, pp.\  259--282. 1987.

\bibitem[Matena \& Raffel(2022)Matena and Raffel]{matena2022merging}
Matena, M.~S. and Raffel, C.~A.
\newblock Merging models with fisher-weighted averaging.
\newblock \emph{Advances in Neural Information Processing Systems}, 35:\penalty0 17703--17716, 2022.

\bibitem[McMahan et~al.(2017)McMahan, Moore, Ramage, Hampson, and y~Arcas]{mcmahan2017communication}
McMahan, B., Moore, E., Ramage, D., Hampson, S., and y~Arcas, B.~A.
\newblock Communication-efficient learning of deep networks from decentralized data.
\newblock In \emph{Artificial intelligence and statistics}, pp.\  1273--1282, 2017.

\bibitem[Meng et~al.(2019)Meng, Zheng, Zhang, Chen, Ye, Ma, Yu, and Liu]{meng2019g}
Meng, Q., Zheng, S., Zhang, H., Chen, W., Ye, Q., Ma, Z.-M., Yu, N., and Liu, T.-Y.
\newblock G-sgd: Optimizing relu neural networks in its positively scale-invariant space.
\newblock In \emph{International Conference on Learning Representations}, 2019.

\bibitem[Nair \& Hinton(2010)Nair and Hinton]{nair2010rectified}
Nair, V. and Hinton, G.~E.
\newblock Rectified linear units improve restricted boltzmann machines.
\newblock In \emph{International Conference on Machine Learning}, pp.\  807--814, 2010.

\bibitem[Navon et~al.(2023)Navon, Shamsian, Achituve, Fetaya, Chechik, and Maron]{navon2023equivariant}
Navon, A., Shamsian, A., Achituve, I., Fetaya, E., Chechik, G., and Maron, H.
\newblock Equivariant architectures for learning in deep weight spaces.
\newblock In \emph{International Conference on Machine Learning}, pp.\  25790--25816, 2023.

\bibitem[Navon et~al.(2024)Navon, Shamsian, Fetaya, Chechik, Dym, and Maron]{navon2024equivariant}
Navon, A., Shamsian, A., Fetaya, E., Chechik, G., Dym, N., and Maron, H.
\newblock Equivariant deep weight space alignment.
\newblock In \emph{International Conference on Machine Learning}, pp.\  37376--37395, 2024.

\bibitem[Neyshabur et~al.(2015)Neyshabur, Salakhutdinov, and Srebro]{neyshabur2015path}
Neyshabur, B., Salakhutdinov, R.~R., and Srebro, N.
\newblock Path-sgd: Path-normalized optimization in deep neural networks.
\newblock \emph{Advances in neural information processing systems}, 28, 2015.

\bibitem[Park \& Kim(2022)Park and Kim]{park2022visiontransformerswork}
Park, N. and Kim, S.
\newblock How do vision transformers work?, 2022.
\newblock URL \url{https://arxiv.org/abs/2202.06709}.

\bibitem[Pe{\~n}a et~al.(2023)Pe{\~n}a, Medeiros, Dubail, Aminbeidokhti, Granger, and Pedersoli]{pena2023re}
Pe{\~n}a, F. A.~G., Medeiros, H.~R., Dubail, T., Aminbeidokhti, M., Granger, E., and Pedersoli, M.
\newblock Re-basin via implicit sinkhorn differentiation.
\newblock In \emph{Proceedings of the IEEE/CVF Conference on Computer Vision and Pattern Recognition}, pp.\  20237--20246, 2023.

\bibitem[Radford et~al.(2021)Radford, Kim, Hallacy, Ramesh, Goh, Agarwal, Sastry, Askell, Mishkin, Clark, et~al.]{radford2021learning}
Radford, A., Kim, J.~W., Hallacy, C., Ramesh, A., Goh, G., Agarwal, S., Sastry, G., Askell, A., Mishkin, P., Clark, J., et~al.
\newblock Learning transferable visual models from natural language supervision.
\newblock In \emph{International conference on machine learning}, pp.\  8748--8763, 2021.

\bibitem[Raffel et~al.(2020)Raffel, Shazeer, Roberts, Lee, Narang, Matena, Zhou, Li, and Liu]{raffel2020exploring}
Raffel, C., Shazeer, N., Roberts, A., Lee, K., Narang, S., Matena, M., Zhou, Y., Li, W., and Liu, P.~J.
\newblock Exploring the limits of transfer learning with a unified text-to-text transformer.
\newblock \emph{Journal of machine learning research}, 21\penalty0 (140):\penalty0 1--67, 2020.

\bibitem[Rame et~al.(2022)Rame, Kirchmeyer, Rahier, Rakotomamonjy, Gallinari, and Cord]{rame2022diverse}
Rame, A., Kirchmeyer, M., Rahier, T., Rakotomamonjy, A., Gallinari, P., and Cord, M.
\newblock Diverse weight averaging for out-of-distribution generalization.
\newblock \emph{Advances in Neural Information Processing Systems}, 35:\penalty0 10821--10836, 2022.

\bibitem[Rossi et~al.(2023)Rossi, Singh, and Hannagan]{rossi2023permutation}
Rossi, S., Singh, A., and Hannagan, T.
\newblock On permutation symmetries in bayesian neural network posteriors: a variational perspective.
\newblock \emph{Advances in Neural Information Processing Systems}, 36, 2023.

\bibitem[Sagi \& Rokach(2018)Sagi and Rokach]{sagi2018ensemble}
Sagi, O. and Rokach, L.
\newblock Ensemble learning: A survey.
\newblock \emph{Wiley interdisciplinary reviews: data mining and knowledge discovery}, 8\penalty0 (4):\penalty0 e1249, 2018.

\bibitem[Sch{\"o}nemann(1966)]{schonemann1966generalized}
Sch{\"o}nemann, P.~H.
\newblock A generalized solution of the orthogonal procrustes problem.
\newblock \emph{Psychometrika}, 31\penalty0 (1):\penalty0 1--10, 1966.

\bibitem[Simsek et~al.(2021)Simsek, Ged, Jacot, Spadaro, Hongler, Gerstner, and Brea]{simsek2021geometry}
Simsek, B., Ged, F., Jacot, A., Spadaro, F., Hongler, C., Gerstner, W., and Brea, J.
\newblock Geometry of the loss landscape in overparameterized neural networks: Symmetries and invariances.
\newblock In \emph{International Conference on Machine Learning}, pp.\  9722--9732, 2021.

\bibitem[Singh \& Jaggi(2020)Singh and Jaggi]{singh2020model}
Singh, S.~P. and Jaggi, M.
\newblock Model fusion via optimal transport.
\newblock \emph{Advances in Neural Information Processing Systems}, 33:\penalty0 22045--22055, 2020.

\bibitem[Strang(1976)]{strang1976linear}
Strang, G.
\newblock \emph{Linear algebra and its applications}.
\newblock Academic Press, 1976.

\bibitem[Tatro et~al.(2020)Tatro, Chen, Das, Melnyk, Sattigeri, and Lai]{tatro2020optimizing}
Tatro, N., Chen, P.-Y., Das, P., Melnyk, I., Sattigeri, P., and Lai, R.
\newblock Optimizing mode connectivity via neuron alignment.
\newblock \emph{Advances in Neural Information Processing Systems}, 33:\penalty0 15300--15311, 2020.

\bibitem[Tjong Kim~Sang \& De~Meulder(2003)Tjong Kim~Sang and De~Meulder]{tjong-kim-sang-de-meulder-2003-introduction}
Tjong Kim~Sang, E.~F. and De~Meulder, F.
\newblock Introduction to the {C}o{NLL}-2003 shared task: Language-independent named entity recognition.
\newblock In \emph{Proceedings of the Seventh Conference on Natural Language Learning at {HLT}-{NAACL} 2003}, pp.\  142--147, 2003.
\newblock URL \url{https://aclanthology.org/W03-0419}.

\bibitem[Tran et~al.(2024)Tran, Vo, The, Huu, Nguyen-Nhat, Tran, Pham, and Nguyen]{tran2024equivariant}
Tran, V.-H., Vo, T.~N., The, A.~N., Huu, T.~T., Nguyen-Nhat, M.-K., Tran, T., Pham, D.-T., and Nguyen, T.~M.
\newblock Equivariant neural functional networks for transformers.
\newblock \emph{arXiv preprint arXiv:2410.04209}, 2024.

\bibitem[Umeyama(1991)]{umeyama1991least}
Umeyama, S.
\newblock Least-squares estimation of transformation parameters between two point patterns.
\newblock \emph{IEEE Transactions on Pattern Analysis \& Machine Intelligence}, 13\penalty0 (04):\penalty0 376--380, 1991.

\bibitem[Vaswani et~al.(2017)Vaswani, Shazeer, Parmar, Uszkoreit, Jones, Gomez, Kaiser, and Polosukhin]{vaswani2017attention}
Vaswani, A., Shazeer, N., Parmar, N., Uszkoreit, J., Jones, L., Gomez, A.~N., Kaiser, {\L}., and Polosukhin, I.
\newblock Attention is all you need.
\newblock \emph{Advances in neural information processing systems}, 30, 2017.

\bibitem[Wang et~al.(2019)Wang, Singh, Michael, Hill, Levy, and Bowman]{wang2019glue}
Wang, A., Singh, A., Michael, J., Hill, F., Levy, O., and Bowman, S.~R.
\newblock Glue: A multi-task benchmark and analysis platform for natural language understanding, 2019.

\bibitem[Wang et~al.(2020)Wang, Yurochkin, Sun, Papailiopoulos, and Khazaeni]{wang2020federated}
Wang, H., Yurochkin, M., Sun, Y., Papailiopoulos, D., and Khazaeni, Y.
\newblock Federated learning with matched averaging.
\newblock In \emph{International Conference on Learning Representations}, 2020.

\bibitem[Wei \& Liu(2025)Wei and Liu]{wei2025large}
Wei, X. and Liu, L.
\newblock Are large language models good in-context learners for financial sentiment analysis?
\newblock \emph{arXiv preprint arXiv:2503.04873}, 2025.

\bibitem[Wortsman et~al.(2022)Wortsman, Ilharco, Gadre, Roelofs, Gontijo-Lopes, Morcos, Namkoong, Farhadi, Carmon, Kornblith, et~al.]{wortsman2022model}
Wortsman, M., Ilharco, G., Gadre, S.~Y., Roelofs, R., Gontijo-Lopes, R., Morcos, A.~S., Namkoong, H., Farhadi, A., Carmon, Y., Kornblith, S., et~al.
\newblock Model soups: averaging weights of multiple fine-tuned models improves accuracy without increasing inference time.
\newblock In \emph{International conference on machine learning}, pp.\  23965--23998, 2022.

\bibitem[Yadav et~al.(2023)Yadav, Tam, Choshen, Raffel, and Bansal]{yadav2023ties}
Yadav, P., Tam, D., Choshen, L., Raffel, C.~A., and Bansal, M.
\newblock Ties-merging: Resolving interference when merging models.
\newblock \emph{Advances in Neural Information Processing Systems}, 36, 2023.

\bibitem[Yadav et~al.(2024)Yadav, Vu, Lai, Chronopoulou, Faruqui, Bansal, and Munkhdalai]{yadav2024matters}
Yadav, P., Vu, T., Lai, J., Chronopoulou, A., Faruqui, M., Bansal, M., and Munkhdalai, T.
\newblock What matters for model merging at scale?
\newblock \emph{arXiv preprint arXiv:2410.03617}, 2024.

\bibitem[Yang et~al.(2024)Yang, Wang, Shen, Liu, Guo, Wang, and Tao]{yang2024adamerging}
Yang, E., Wang, Z., Shen, L., Liu, S., Guo, G., Wang, X., and Tao, D.
\newblock Adamerging: Adaptive model merging for multi-task learning.
\newblock In \emph{International Conference on Learning Representations}, 2024.

\bibitem[Yun et~al.(2019)Yun, Jeong, Kim, Kang, and Kim]{yun2019graph}
Yun, S., Jeong, M., Kim, R., Kang, J., and Kim, H.~J.
\newblock Graph transformer networks.
\newblock \emph{Advances in neural information processing systems}, 32, 2019.

\bibitem[Yurochkin et~al.(2019)Yurochkin, Agarwal, Ghosh, Greenewald, Hoang, and Khazaeni]{yurochkin2019bayesian}
Yurochkin, M., Agarwal, M., Ghosh, S., Greenewald, K., Hoang, N., and Khazaeni, Y.
\newblock Bayesian nonparametric federated learning of neural networks.
\newblock In \emph{International conference on machine learning}, pp.\  7252--7261, 2019.

\bibitem[Zamir et~al.(2025)Zamir, Dokania, Zhao, and Yu]{zamir2025improving}
Zamir, G., Dokania, A., Zhao, B., and Yu, R.
\newblock Improving learning to optimize using parameter symmetries.
\newblock \emph{arXiv preprint arXiv:2504.15399}, 2025.

\bibitem[Zhang et~al.(2025)Zhang, Chen, Zheng, Li, and Chen]{zhang2025resolving}
Zhang, B., Chen, Z., Zheng, Z., Li, J., and Chen, H.
\newblock Resolving editing-unlearning conflicts: A knowledge codebook framework for large language model updating.
\newblock \emph{arXiv preprint arXiv:2502.00158}, 2025.

\bibitem[Zhao et~al.(2022)Zhao, Dehmamy, Walters, and Yu]{zhao2022symmetry}
Zhao, B., Dehmamy, N., Walters, R., and Yu, R.
\newblock Symmetry teleportation for accelerated optimization.
\newblock \emph{Advances in neural information processing systems}, 35:\penalty0 16679--16690, 2022.

\bibitem[Zhao et~al.(2023)Zhao, Ganev, Walters, Yu, and Dehmamy]{zhao2023symmetries}
Zhao, B., Ganev, I., Walters, R., Yu, R., and Dehmamy, N.
\newblock Symmetries, flat minima, and the conserved quantities of gradient flow.
\newblock In \emph{The Eleventh International Conference on Learning Representations}, 2023.

\bibitem[Zhao et~al.(2024)Zhao, Gower, Walters, and Yu]{zhao2024improving}
Zhao, B., Gower, R.~M., Walters, R., and Yu, R.
\newblock Improving convergence and generalization using parameter symmetries.
\newblock In \emph{International Conference on Learning Representations}, 2024.

\bibitem[Zheng et~al.(2024)Zheng, Dong, Wang, Liu, Wang, and Li]{zheng2024kg}
Zheng, Z., Dong, Y., Wang, S., Liu, H., Wang, Q., and Li, J.
\newblock Kg-cf: Knowledge graph completion with context filtering under the guidance of large language models.
\newblock In \emph{2024 IEEE International Conference on Big Data (BigData)}, pp.\  805--810, 2024.

\bibitem[Zhou et~al.(2023{\natexlab{a}})Zhou, Yang, Burns, Cardace, Jiang, Sokota, Kolter, and Finn]{zhou2023permutation}
Zhou, A., Yang, K., Burns, K., Cardace, A., Jiang, Y., Sokota, S., Kolter, J.~Z., and Finn, C.
\newblock Permutation equivariant neural functionals.
\newblock \emph{Advances in neural information processing systems}, 36:\penalty0 24966--24992, 2023{\natexlab{a}}.

\bibitem[Zhou et~al.(2023{\natexlab{b}})Zhou, Yang, Jiang, Burns, Xu, Sokota, Kolter, and Finn]{zhou2023neural}
Zhou, A., Yang, K., Jiang, Y., Burns, K., Xu, W., Sokota, S., Kolter, J.~Z., and Finn, C.
\newblock Neural functional transformers.
\newblock \emph{Advances in neural information processing systems}, 36:\penalty0 77485--77502, 2023{\natexlab{b}}.

\bibitem[Zhou et~al.(2021)Zhou, Zhang, Peng, Zhang, Li, Xiong, and Zhang]{zhou2021informer}
Zhou, H., Zhang, S., Peng, J., Zhang, S., Li, J., Xiong, H., and Zhang, W.
\newblock Informer: Beyond efficient transformer for long sequence time-series forecasting.
\newblock In \emph{Proceedings of the AAAI conference on artificial intelligence}, volume~35, pp.\  11106--11115, 2021.

\bibitem[Zhou et~al.(2024)Zhou, Song, Wang, and Chen]{zhou2024metagpt}
Zhou, Y., Song, L., Wang, B., and Chen, W.
\newblock Metagpt: Merging large language models using model exclusive task arithmetic.
\newblock In \emph{Proceedings of the 2024 Conference on Empirical Methods in Natural Language Processing}, pp.\  1711--1724, 2024.

\bibitem[Zhou et~al.(2023{\natexlab{c}})Zhou, Yang, Yang, Yan, and Hu]{zhou2023going}
Zhou, Z., Yang, Y., Yang, X., Yan, J., and Hu, W.
\newblock Going beyond linear mode connectivity: The layerwise linear feature connectivity.
\newblock \emph{Advances in Neural Information Processing Systems}, 36, 2023{\natexlab{c}}.

\bibitem[Zhu et~al.(2024)Zhu, Wu, Zhang, Wang, Guo, Hong, Simon, and Li]{zhu2024understanding}
Zhu, Y., Wu, L., Zhang, B., Wang, S., Guo, Q., Hong, L., Simon, L., and Li, J.
\newblock Understanding and modeling job marketplace with pretrained language models.
\newblock In \emph{Proceedings of the 33rd ACM International Conference on Information and Knowledge Management}, pp.\  5143--5150, 2024.

\bibitem[Ziyin et~al.(2024)Ziyin, Wang, Li, and Wu]{ziyin2024parameter}
Ziyin, L., Wang, M., Li, H., and Wu, L.
\newblock Parameter symmetry and noise equilibrium of stochastic gradient descent.
\newblock \emph{arXiv preprint arXiv:2402.07193}, 2024.

\bibitem[Ziyin et~al.(2025)Ziyin, Xu, and Chuang]{ziyin2025remove}
Ziyin, L., Xu, Y., and Chuang, I.~L.
\newblock Remove symmetries to control model expressivity and improve optimization.
\newblock In \emph{International Conference on Learning Representations}, 2025.

\end{thebibliography}
\bibliographystyle{icml2025}

\newpage
\appendix
\onecolumn
\section{Proof}\label{sec:appendix_proof}
\paragraph{\Cref{thm:solution}.}
\textit{The following optimization problem has a closed-form solution.}
\begin{equation}
\begin{aligned}
\min_{\bm{R}_1,\bm{R}_2\in\mathcal{R}}&\left\|\left[
\begin{array}{cc}
\bm{W}_{Q_1}^\top & \bm{W}_{Q_2}^\top \\
\bm{b}_{Q_1} & \bm{b}_{Q_2} \\
\end{array}\right]\left[
\begin{array}{c}
\bm{R}_1 \\
-\bm{R}_2 \\
\end{array}\right]\right\|_F^2+\left\|\left[
\begin{array}{cc}
\bm{W}_{K_1}^\top & \bm{W}_{K_2}^\top \\
\bm{b}_{K_1} & \bm{b}_{K_2} \\
\end{array}\right]\left[
\begin{array}{c}
\bm{R}_1 \\
-\bm{R}_2 \\
\end{array}\right]\right\|_F^2.
\end{aligned}
\end{equation}
\textit{The solution is given by}
\begin{equation}
\bm{R}_1=\bm{U}\bm{V}^\top,\bm{R}_2=\bm{I},
\end{equation}
\textit{where $\bm{I}$ is the identity matrix and $\bm{U}\bm{\Sigma}\bm{V}^\top=\bm{W}_{Q_1}\bm{W}_{Q_2}^\top+\bm{W}_{K_1}\bm{W}_{K_2}^\top+\bm{b}_{Q_1}^\top\bm{b}_{Q_2}+\bm{b}_{K_1}^\top\bm{b}_{K_2}$ is the result of eigendecomposition.}

\begin{proof}
We first show the process of converting the optimization problem to an Orthogonal Procrustes problem~\citep{schonemann1966generalized,gower2004procrustes}.
We can obtain that the optimization problem is equivalent to the following one.
\begin{equation}\label{eq:objective}
\begin{aligned}
\min_{\bm{R}_1,\bm{R}_2\in\mathcal{R}}&\left\|\bm{W}_{Q_1}^\top\bm{R}_1-\bm{W}_{Q_2}^\top\bm{R}_2\right\|_F^2+\left\|\bm{b}_{Q_1}\bm{R}_1-\bm{b}_{Q_2}\bm{R}_2\right\|_F^2+\left\|\bm{W}_{K_1}^\top\bm{R}_1-\bm{W}_{K_2}^\top\bm{R}_2\right\|_F^2+\left\|\bm{b}_{K_1}\bm{R}_1-\bm{b}_{K_2}\bm{R}_2\right\|_F^2.
\end{aligned}
\end{equation}
Considering that $\bm{R}_1,\bm{R}_2\in\mathcal{R}$ and $\bm{R}_1,\bm{R}_2$ are nonsingular matrices, we let $\bm{R}=\bm{R}_1\bm{R}_2^{-1}$ and convert \Cref{eq:objective} to the following one:
\begin{equation}\label{eq:objective1}
\begin{aligned}
\min_{\bm{R},\bm{R}_2\in\mathcal{R}}&\left\|\left(\bm{W}_{Q_1}^\top\bm{R}-\bm{W}_{Q_2}^\top\right)\bm{R}_2\right\|_F^2+\left\|\left(\bm{b}_{Q_1}\bm{R}-\bm{b}_{Q_2}\right)\bm{R}_2\right\|_F^2+\left\|\left(\bm{W}_{K_1}^\top\bm{R}-\bm{W}_{K_2}^\top\right)\bm{R}_2\right\|_F^2+\left\|\left(\bm{b}_{K_1}\bm{R}-\bm{b}_{K_2}\right)\bm{R}_2\right\|_F^2.
\end{aligned}
\end{equation}
As multiplying $\bm{R}_2$ preserves the Frobenius norm, we can remove the $\bm{R}_2$ terms from the objective and obtain an Orthogonal Procrustes problem as follows:
\begin{equation}\label{eq:objective2}
\begin{aligned}
\min_{\bm{R}\in\mathcal{R}}&\left\|\bm{W}_{Q_1}^\top\bm{R}-\bm{W}_{Q_2}^\top\right\|_F^2+\left\|\bm{b}_{Q_1}\bm{R}-\bm{b}_{Q_2}\right\|_F^2+\left\|\bm{W}_{K_1}^\top\bm{R}-\bm{W}_{K_2}^\top\right\|_F^2+\left\|\bm{b}_{K_1}\bm{R}-\bm{b}_{K_2}\right\|_F^2.
\end{aligned}
\end{equation}
We take a look at the first term $\left\|\bm{W}_{Q_1}^\top\bm{R}-\bm{W}_{Q_2}^\top\right\|_F^2$ and have:
\begin{equation}
\begin{aligned}
&\min_{\bm{R}\in\mathcal{R}}\left\|\bm{W}_{Q_1}^\top\bm{R}-\bm{W}_{Q_2}^\top\right\|_F^2 \\
=&\min_{\bm{R}\in\mathcal{R}}\left<\bm{W}_{Q_1}^\top\bm{R}-\bm{W}_{Q_2}^\top,\bm{W}_{Q_1}^\top\bm{R}-\bm{W}_{Q_2}^\top\right>_F \\
=&\max_{\bm{R}\in\mathcal{R}}\left<\bm{W}_{Q_1}^\top\bm{R},\bm{W}_{Q_2}^\top\right>_F \\
=&\max_{\bm{R}\in\mathcal{R}}\mathrm{tr}\left(\bm{R}^\top\bm{W}_{Q_1}\bm{W}_{Q_2}^\top\right) \\
=&\max_{\bm{R}\in\mathcal{R}}\left<\bm{R},\bm{W}_{Q_1}\bm{W}_{Q_2}^\top\right>_F. \\
\end{aligned}
\end{equation}
Similarly, we can convert \Cref{eq:objective2} to the following one:
\begin{equation}
\max_{\bm{R}\in\mathcal{R}}\left<\bm{R},\bm{W}_{Q_1}\bm{W}_{Q_2}^\top+\bm{W}_{K_1}\bm{W}_{K_2}^\top+\bm{b}_{Q_1}^\top\bm{b}_{Q_2}+\bm{b}_{K_1}^\top\bm{b}_{K_2}\right>_F.
\end{equation}
We then conduct the singular value decomposition to the matrix $\bm{W}_{Q_1}\bm{W}_{Q_2}^\top+\bm{W}_{K_1}\bm{W}_{K_2}^\top+\bm{b}_{Q_1}^\top\bm{b}_{Q_2}+\bm{b}_{K_1}^\top\bm{b}_{K_2}=\bm{U}\bm{\Sigma}\bm{V}^\top$ and have:
\begin{equation}
\begin{aligned}
&\max_{\bm{R}\in\mathcal{R}}\left<\bm{R},\bm{W}_{Q_1}\bm{W}_{Q_2}^\top+\bm{W}_{K_1}\bm{W}_{K_2}^\top+\bm{b}_{Q_1}^\top\bm{b}_{Q_2}+\bm{b}_{K_1}^\top\bm{b}_{K_2}\right>_F \\
=&\max_{\bm{R}\in\mathcal{R}}\mathrm{tr}\left(\bm{V}\bm{\Sigma}\bm{U}^\top\bm{R}\right) \\
=&\max_{\bm{R}\in\mathcal{R}}\mathrm{tr}\left(\bm{\Sigma}\bm{U}^\top\bm{R}\bm{V}\right) \\
=&\mathrm{tr}(\bm{\Sigma}),
\end{aligned}
\end{equation}
where $\bm{U}^\top\bm{R}\bm{V}=\bm{I}$, \textit{i.e.}, $\bm{R}=\bm{U}\bm{V}^\top$.
Finally, we incorporate $\bm{R}=\bm{U}\bm{V}^\top$ into $\bm{R}=\bm{R}_1\bm{R}_2^{-1}$ and let $\bm{R}_2=\bm{I}$ for simplicity.
Subsequently, we have $\bm{R}_1=\bm{U}\bm{V}^\top$ and $\bm{R}_2=\bm{I}$ as a closed-form solution of the optimization problem in \Cref{thm:solution}.

\end{proof}

From the proof, we can observe that the optimization problem in \Cref{thm:solution} has infinite pairs of solutions $\{\bm{R}_1,\bm{R}_2\}$ regarding the value of $\bm{R}_2$.
In this paper, we let $\bm{R}_2=\bm{I}$ for simplicity, which makes model 2 an anchor model.
However, the value of $\bm{R}_2$ can be further adjusted to benefit some data-dependent model fusion techniques~\citep{singh2020model,jin2023dataless}.
We leave this part to future work.

\section{Comparison with Related Work}

Git Re-Basin~\citep{ainsworth2023git} introduces three model fusion algorithms: weight matching, activation matching, and straight-through matching, based on permutation symmetry. 
While these methods are effective for CNNs and MLPs, they do not easily extend to transformers due to the discrete nature of permutation symmetry. 
In a discrete space, the parameter matching problem is equivalent to solving a sum of bilinear assignment problems, which is NP-hard~\citep{ainsworth2023git}.
To address this limitation, we introduce rotation symmetry, which extends the symmetry space to a continuous domain, allowing for a closed-form solution to parameter matching. 
Unlike permutation symmetry, rotation symmetry provides a more flexible and efficient solution for aligning parameters in transformers. 

Additionally, our study emphasizes the advantage of weight matching as a plug-and-play module for model fusion, as it preserves model functionality while reducing the inner distance of end models after matching. 
OT-ACTS~\citep{imfeld2024transformer} also includes a parameter matching step, leveraging optimal transport for model fusion. 
However, its matching module is restricted to permutation operations, and its fusion method is limited to simple merging. 
In contrast, our approach generalizes parameter matching to rotation operations and integrates it with more advanced merging strategies, demonstrating the potential of parameter matching to enhance model fusion.

A concurrent work by~\citet{tran2024equivariant} explores a similar form of rotation symmetry. However, their focus is on constructing functionally equivalent networks for transformers, whereas we focus on leveraging rotation symmetry for model fusion, proposing a theoretically optimal parameter matching algorithm specifically designed to improve fusion performance.




\section{Experimental Details}\label{app:b}

\paragraph{Datasets.}
Similar with~\citep{jin2023dataless}, We employ emotion classification and named entity recognition (NER) as the tasks for the main experiments.
The emotion classification datasets are extracted from~\citep{bostan-klinger-2018-analysis}. 
Five of them are selected as in-domain datasets (emoint, ssec, electoraltweets, grounded\_emotions, affectivetext), and five others are good datasets (dailydialog, crowdflower, tec, tales-emotion, isear). 
For the NER task, we choose OntoNotes~\citep{hovy-etal-2006-ontonotes} for model finetuning and CoNLL~\citep{tjong-kim-sang-de-meulder-2003-introduction} for out-of-domain evaluation.
Additionally, in the ablation study and subset layers matching, we employ datasets (STS-B, SST-2, CoLA) from GLUE~\citep{wang2019glue} to analyze the module performance. 

\paragraph{Baselines.}
We compared the performance of our method with two matching baselines:
\begin{itemize}[itemsep=0pt, parsep=0pt, leftmargin=*, topsep=0pt]
    \item \textbf{OTFusion}~\citep{singh2020model} mitigates the positional mismatch of functionally analogous neurons across end models by employing optimal transport to align neurons layer-wise, thus enabling more effective model merging.
    \item \textbf{Git Re-Basin}~\citep{ainsworth2023git} matches the parameters of end models through learned permutation symmetries in the parameter space, aiming to transform their parameters into a shared basin within the loss landscape and facilitate seamless merging. 
\end{itemize}

Meanwhile, to address RQ1, we further select six merging methods as baselines:
\begin{itemize}[itemsep=0pt, parsep=0pt, leftmargin=*, topsep=0pt]
    \item \textbf{Simple}~\citep{wortsman2022model} directly averages the weight matrices of independently pretrained end models.
    \item \textbf{Fisher}~\citep{matena2022merging} approximates each end model’s posterior as a Gaussian distribution with a precision matrix derived from its Fisher information, thereby casting model merging as a Maximum Likelihood Estimation (MLE) problem over these posterior distributions.
    \item \textbf{RegMean}~\citep{jin2023dataless} formulates model merging as a regression problem, aiming to minimize the Euclidean distance between the predictions of the merged model and individual end models over a set of input samples.
    \item \textbf{OT-WTS}~\citep{imfeld2024transformer} extends OTFusion by introducing soft neuron alignment, which replaces the rigid one-to-one neuron matching with a more flexible, weighted correspondence between neurons.
    Subsequent model merging is then guided by these soft alignment weights.
    \item \textbf{OT-ACTS} and \textbf{OT-ACTS-EMD}~\citep{imfeld2024transformer} are two variants of \textbf{OT-WTS} that match neurons based on their activations rather than directly aligning parameter matrices.
    Specifically, OT-ACTS measures activation similarity using Euclidean distance, while OT-ACTS-EMD employs Earth Mover’s Distance for a more fine-grained comparison.
\end{itemize}

The code implementation is based on~\citep{jin2023dataless}\footnote{\url{https://github.com/bloomberg/dataless-model-merging}} for NLP tasks and~\citep{imfeld2024transformer}\footnote{\url{https://github.com/graldij/transformer-fusion}} for vision tasks.


\section{Limitations}

In this study, we proposed a theoretically optimal parameter matching algorithm based on rotation symmetry. However, the optimality guarantees of our method are subject to several important prerequisites.


First, our theoretical analysis in \Cref{thm:solution} focuses on the binary model matching case, where the optimal solution is independent of the choice of anchor model. This property, however, does not naturally extend to the multi-model setting. When matching multiple models, a straightforward extension involves selecting a single model as an anchor and aligning the others to it using pairwise rotation-based matching. Yet, in this case, the global optimality depends on the definition of the overall distance metric and may not be preserved. An alternative is iterative pairwise matching, but this introduces path dependency, i.e., the final result can vary depending on the order of merging. Extending our framework to the optimal multi-model matching problem remains an open challenge for future work.

Second, the scope of our parameter matching algorithm is within the rotation symmetry, as defined in \Cref{eq:match_attn}. While this serves as a tractable and well-motivated instance of parameter space symmetry, we recognize that Transformers exhibit a richer set of structural symmetries beyond permutation, rotation, and rescaling. For example, recent work suggests that geometric properties of normalization and softmax layers may enlarge the symmetry set of previous model layers~\citep{kunin2021neural,lim2024graph}. Characterizing the complete symmetry group of transformers remains an open question. 

We also considered the possibility of extending rotation matrices to invertible matrices in terms of parameter space symmetries. Although invertible matrices expand the symmetry set, they are less suitable for practical parameter matching. Specifically, our closed-form solution in \Cref{eq:solution} relies on the orthogonality of the rotation matrices. 
If replacing them with general invertible matrices, it becomes difficult to derive a closed-form solution for~\Cref{eq:solution}, leading to suboptimal solution and increased complexity. Additionally, parameter matching requires computation of the inverse matrix of every $\bm{R}$, which is impractical for invertible matrices. 
In contrast, orthogonal matrices benefit from computationally efficient inverses (via transposition) and theoretical simplicity.


Finally, we apply the rescaling symmetry after optimizing for rotation-based matching. However, our sequential approach (first optimizing rotation $\bm{R}$, then rescaling weight $\alpha$) does not guarantee global optimality for the joint optimization over $(\bm{R},\alpha)$. Deriving a globally optimal solution would require solving~\Cref{eq:rescaling_derivative} in closed form, which remains non-trivial. Hence, we consider our method as a practical approximation to the joint optimization solution which balances theoretical soundness with computational efficiency. Exploring a fully joint optimization framework for rotation and rescaling symmetries can be a promising direction for future research.



\end{document}